\documentclass[journal]{IEEEtran}
\usepackage{cite}
\usepackage{amsmath,amssymb,amsfonts}
\usepackage{algorithmic}
\usepackage{graphicx}
\usepackage{textcomp}
\usepackage{rotating}
\usepackage{tikz}
\usepackage{hyperref}
\usepackage{multirow}
\usepackage{booktabs}
\usepackage[normalem]{ulem}
\useunder{\uline}{\ul}{}
\usepackage{lscape}
\usepackage{longtable}

\def\BibTeX{{\rm B\kern-.05em{\sc i\kern-.025em b}\kern-.08em
    T\kern-.1667em\lower.7ex\hbox{E}\kern-.125emX}}
\begin{document}



\bstctlcite{IEEEexample:BSTcontrol}
    \title{Ten Years of Generative Adversarial Nets (GANs): A survey of the state-of-the-art}
  \author{Tanujit Chakraborty, Ujjwal Reddy K S, Shraddha M. Naik, Madhurima Panja, and~Bayapureddy Manvitha
  
\thanks{Corresponding Author: T. Chakraborty is with the Department of Science and Engineering, Sorbonne University Abu Dhabi, UAE and Sorbonne Centre for Artificial Intelligence, Sorbonne Université, Paris. (e-mail: tanujit.chakraborty@sorbonne.ae).}
\thanks{U. Reddy K S and B. Manvitha are with Vellore Institute Of Technology, Andhra Pradesh, India. (e-mail: ujjwalreddyks@gmail.com and manvitha.bayapureddy@gmail.com).}
\thanks{S M. Naik is with the Department of Science and Engineering, Sorbonne University Abu Dhabi, UAE. (e-mail: 7.shraddha.naik@gmail.com).}
\thanks{M. Panja is with the Center for Data Sciences, International Institute of Information Technology Bangalore, India. (e-mail: madhruima.panja@iiitb.ac.in).}
\thanks{All authors have contributed equally to this survey.}}


\maketitle



\begin{abstract}
Since their inception in 2014, Generative Adversarial Networks (GANs) have rapidly emerged as powerful tools for generating realistic and diverse data across various domains, including computer vision and other applied areas. Consisting of a discriminative network and a generative network engaged in a Minimax game, GANs have revolutionized the field of generative modeling. In February 2018, GAN secured the leading spot on the ``Top Ten Global Breakthrough Technologies List'' issued by the Massachusetts Science and Technology Review. Over the years, numerous advancements have been proposed, leading to a rich array of GAN variants, such as conditional GAN, Wasserstein GAN, CycleGAN, and StyleGAN, among many others. This survey aims to provide a general overview of GANs, summarizing the latent architecture, validation metrics, and application areas of the most widely recognized variants. We also delve into recent theoretical developments, exploring the profound connection between the adversarial principle underlying GAN and Jensen-Shannon divergence, while discussing the optimality characteristics of the GAN framework. The efficiency of GAN variants and their model architectures will be evaluated along with training obstacles as well as training solutions. In addition, a detailed discussion will be provided, examining the integration of GANs with newly developed deep learning frameworks such as Transformers, Physics-Informed Neural Networks, Large Language models, and Diffusion models. Finally, we reveal several issues as well as future research outlines in this field.
\end{abstract}




\section{Introduction}\label{Section_Introduction}
\PARstart{G}{enerative} Adversarial Networks (GANs) have emerged as a transformative deep learning approach for generating high-quality and diverse data. In GAN, a generator network produces data, while a discriminator network evaluates the authenticity of the generated data. Through an adversarial mechanism, the discriminator learns to distinguish between real and fake data, while the generator aims to produce data that is indistinguishable from real data.

Since their introduction in 2014 by Goodfellow et al. \cite{goodfellow2014generative}, GANs have witnessed remarkable advancements, leading to the development of numerous specialized variants that excel in creating data across diverse fields. Conditional GAN \cite{mirza2014conditional} enables the generation of data based on specific conditions or desired qualities, such as synthesizing photos of a particular class. CycleGAN \cite{zhu2017unpaired} have proven effective in image-to-image translation tasks, even in the absence of paired data. StackGAN \cite{zhang2017stackgan} has demonstrated the ability to generate high-resolution images from textual descriptions, pushing the boundaries of visual realism. Progressive GAN \cite{karras2017progressive} has achieved exceptional results in producing high-quality images with increasing resolution. StyleGAN \cite{karras2019style}, known for its versatility, generates images with a wide range of styles and distinctive features. Furthermore, GANs have extended beyond visual domains and shown potential in generating textual \cite{liu2018towards}, musical \cite{yang2017midinet}, 3D modeling \cite{wu2016google}, future cities \cite{thottolil2023prediction}, time series \cite{smith2020conditional} data among many others.

The success of GANs has led to their adoption in various applications, such as image and video synthesis, data augmentation, super-resolution, inpainting, anomaly detection, and image editing. GANs have also been employed to address data scarcity issues in machine learning, where they generate synthetic data to improve the effectiveness of models trained on limited datasets \cite{shin2016deep}. Additionally, GANs have found utility in creating realistic simulations for video games and virtual reality environments, enhancing user experiences and immersive interactions \cite{togelius2014procedural}. To ensure the comprehensiveness of this survey, we conducted an extensive review of the research papers encompassing both theoretical advancements and practical applications of GAN. Our survey draws insights from diverse fields, including computer vision, natural language processing, autonomous vehicles, time series, medical domain, and many others. Notable papers that significantly contributed to our survey include Goodfellow et al. \cite{goodfellow2014generative} for introducing the GAN framework, Mirza and Osindero \cite{mirza2014conditional} for pioneering conditional GAN, Zhu et al. \cite{zhu2017unpaired} for introducing CycleGAN, Karras et al. \cite{karras2017progressive} for their seminal work on progressive GAN, and Chen et al. \cite{chen2016infogan} for the breakthroughs achieved with InfoGAN, among many others.

Despite their remarkable achievements, GANs face several challenges in practice. One prominent issue is the instability of the training process, which can result in mode collapse or oscillation \cite{arjovsky2017towards}. Another challenge lies in the evaluation of generated data, as conventional assessment criteria may not adequately capture the diversity and realism of the synthesized samples \cite{wilby2019using}. Furthermore, GANs have been observed to exhibit biases, particularly concerning gender and race, potentially reflecting the biases present in the training data \cite{buolamwini2018gender, zhao2018gender}. To overcome the limitations of GAN various modified training approaches and hybridization with popular deep learning architectures such as Transformers \cite{vaswani2017attention}, Physics-Informed Neural Network (PINN) \cite{raissi2019physics}, Large language models (LLMs) \cite{radford2019better}, and Diffusion models \cite{sohl2015deep} have been proposed in the literature. These modified methodologies have shown promise in enhancing the synthetic data generation capabilities of GANs.

Finally, GANs have emerged as an effective tool for producing high-quality and varied data in several disciplines. Notwithstanding the difficulties connected with their use, GANs have shown outstanding results and have the potential to drive innovation in disciplines such as computer vision, machine learning, and virtual reality. This in-depth analysis covers the accomplishments and limitations of GAN, as well as the promise of these approaches for future research and applications. This comprehensive survey aims to explore both the accomplishments and challenges of GAN. The contributions of the article can be summarized as follows:
\begin{itemize}
    \item \textbf{Exploration of Vanilla GAN and their applications:} We offer an elaborate description of the GAN model, encompassing its architectural particulars and the mathematical optimization functions it employs. We summarize the areas where GANs have emerged as a promising tool in efficiently solving real-world problems with their generative capabilities. 


    \item \textbf{Evolution of state-of-the-art GAN models across the decade:} Our comprehensive analysis encompasses a wide range of cutting-edge GAN adaptations crafted to address practical challenges across various domains. We delve into their structural designs, practical uses, execution methods, and constraints. To facilitate a lucid understanding of the field's progress, we present an intricate chronological breakdown of GAN model advancements. Furthermore, we evaluate recent field surveys, outlining their pros and cons, while also tackling these aspects within our own survey.

    \item \textbf{Theoretical advancements of GANs:} We give a technical overview of the theoretical developments of GANs by exploring the connections between adversarial training and Jensen-Shannon divergence and discussing their optimality features.

    \item \textbf{Assessment of GAN Models:} We provide a comprehensive breakdown of the essential performance measures utilized to assess both the caliber and range of samples produced by GANs. These metrics notably fluctuate depending on the specific domains of application.

    \item \textbf{Limitations of GANs:} We critically examine the constraints associated with GANs, primarily stemming from issues of learning instability, and discuss various enhancement strategies aimed at alleviating these challenges.
    
    \item \textbf{Anticipating future trajectories:} In addition to evaluating the pros and cons of current GAN-centric approaches, we illuminate the hybridization of emerging deep learning models such as Transformers, PINNs, LLMs, and Diffusion models with GANs. We outline potential avenues for research within this domain by summarizing several open scientific problems.  
\end{itemize}


This survey is structured in the following manner. Section \ref{Section_Related_Works} digs into related works and recent surveys giving background information and emphasizing the most significant developments in GAN done over the decade. Section \ref{Section_Overview_GAN} is a concise overview of GAN describing the fundamental components and intricate details of its architecture. In Section \ref{Section_Application}, we examine the wide range of fields that GANs have influenced, such as computer vision, natural language processing, time series, and audio, among many others. Subsequently, Section \ref{gan_vARIANTS_SECTION} reviews the innovations and applications of popular GAN-based frameworks from various domains along with their implementation software and discusses their limitations. This section also provides a timeline for the GAN models to have a clear vision of the development of this field. Section \ref{Section_Theory} summarizes the recent theoretical developments of GAN and its variants. Section \ref{Evaluation} reviews the metrics used for evaluating GAN-based models. Section \ref{Section_Limitations} analyzes the limitations of GANs and presents its remedial measures. Section \ref{Section_Discussion} discusses the potential and usability of GAN with the development of new deep learning technologies such as Transformers, PINNs, LLMs, and Diffusion models. Section \ref{Section_Future} proposes potential directions for further research in this field. Finally, Section \ref{Section_Conclusion} concludes the survey by indicating prospective directions for future research projects while also offering a closing assessment of the successes and limits of GANs.

\section{Related Works and Recent Surveys}\label{Section_Related_Works}
GANs are a promising deep learning framework for generating artificial data that closely resembles real-world data \cite{goodfellow2014generative}. Early GAN-related research focused on creating realistic visuals. Radford et al. proposed a deep convolutional GAN (DCGAN) in 2015 \cite{radford2015unsupervised}, which utilized convolutional layers, batch normalization, and a specific loss function to generate high-quality images. DCGAN introduced important innovations in image generation. In 2017, Karras et al. \cite{karras2017progressive} introduced progressive growing GAN (ProGAN), which generates higher quality and resolution images compared to vanilla GAN. ProGAN trains multiple generators and discriminators in a stepwise manner, gradually increasing the resolution of the generated images. The results demonstrated the ability of ProGAN to produce images closely resembling genuine photos for various datasets, including the CelebA dataset \cite{zhang2020celeba}.

\begin{table*}[!ht]
\caption{Comparison of our survey and other related GAN surveys (green circle signifies ``Fully covered'', blue circle signifies ``Partially covered'', and red circle signifies ``Not covered'').}
  \centering \scriptsize
  \begin{tabular}{|c|c|c|c|c|c|c|c|c|c|c|}
         \hline
        \multirow{3}{*}{Year} & \multirow{3}{*}{Survey} & \multirow{2}{*}{Theoretical} & \multirow{2}{*}{Evaluation} & \multicolumn{7}{c|}{Domain} \\
        \cline{5-11}&&&& Computer & Natural Language & \multirow{2}{*}{Music} & \multirow{2}{*}{Medical} & Time & Urban & Imbalanced\\
        && Insights & Metrics & Vision & Processing & & & Series & Planning & Classification\\ \hline
        2019 & Kulkarni et al. \cite{kulkarni2019survey} & \tikz\draw[blue,fill=blue] (0,0) circle (.7ex); & \tikz\draw[red,fill=red] (0,0) circle (.7ex); & \tikz\draw[red,fill=red] (0,0) circle (.7ex); & \tikz\draw[red,fill=red] (0,0) circle (.7ex); & \tikz\draw[green,fill=green] (0,0) circle (.7ex); & \tikz\draw[red,fill=red] (0,0) circle (.7ex); & \tikz\draw[red,fill=red] (0,0) circle (.7ex); & \tikz\draw[red,fill=red] (0,0) circle (.7ex); & \tikz\draw[red,fill=red] (0,0) circle (.7ex); \\
        
        2021 & Jabbar et al. \cite{jabbar2021survey} & \tikz\draw[green,fill=green] (0,0) circle (.7ex); & 
        \tikz\draw[red,fill=red] (0,0) circle (.7ex); & \tikz\draw[green,fill=green] (0,0) circle (.7ex); & \tikz\draw[green,fill=green] (0,0) circle (.7ex); & \tikz\draw[green,fill=green] (0,0) circle (.7ex); & \tikz\draw[green,fill=green] (0,0) circle (.7ex); & \tikz\draw[red,fill=red] (0,0) circle (.7ex); &  \tikz\draw[red,fill=red] (0,0) circle (.7ex); & \tikz\draw[red,fill=red] (0,0) circle (.7ex); \\ 

        2021 & Durgadevi et al. \cite{durgadevi2021generative} & \tikz\draw[blue,fill=blue] (0,0) circle (.7ex); & \tikz\draw[red,fill=red] (0,0) circle (.7ex); & \tikz\draw[blue,fill=blue] (0,0) circle (.7ex); & \tikz\draw[blue,fill=blue] (0,0) circle (.7ex); & \tikz\draw[red,fill=red] (0,0) circle (.7ex); & \tikz\draw[blue,fill=blue] (0,0) circle (.7ex); & \tikz\draw[red,fill=red] (0,0) circle (.7ex); & \tikz\draw[red,fill=red] (0,0) circle (.7ex); & \tikz\draw[red,fill=red] (0,0) circle (.7ex); \\

        2021 &  Nandhini et al. \cite{nandhini2021deep} & \tikz\draw[red,fill=red] (0,0) circle (.7ex); & \tikz\draw[red,fill=red] (0,0) circle (.7ex); & \tikz\draw[green,fill=green] (0,0) circle (.7ex); & \tikz\draw[red,fill=red] (0,0) circle (.7ex); & \tikz\draw[red,fill=red] (0,0) circle (.7ex); & \tikz\draw[red,fill=red] (0,0) circle (.7ex); & \tikz\draw[red,fill=red] (0,0) circle (.7ex); & \tikz\draw[red,fill=red] (0,0) circle (.7ex); & \tikz\draw[red,fill=red] (0,0) circle (.7ex); \\

        2021 & Wang et al. \cite{wang2021generative} & \tikz\draw[blue,fill=blue] (0,0) circle (.7ex); & \tikz\draw[red,fill=red] (0,0) circle (.7ex); & \tikz\draw[green,fill=green] (0,0) circle (.7ex); & \tikz\draw[red,fill=red] (0,0) circle (.7ex); & \tikz\draw[red,fill=red] (0,0) circle (.7ex); & \tikz\draw[red,fill=red] (0,0) circle (.7ex); & \tikz\draw[red,fill=red] (0,0) circle (.7ex); & \tikz\draw[red,fill=red] (0,0) circle (.7ex); & \tikz\draw[red,fill=red] (0,0) circle (.7ex); \\

        2021 & Sampath et al \cite{sampath2021survey} & \tikz\draw[blue,fill=blue] (0,0) circle (.7ex); & \tikz\draw[red,fill=red] (0,0) circle (.7ex); & \tikz\draw[blue,fill=blue] (0,0) circle (.7ex); & \tikz\draw[red,fill=red] (0,0) circle (.7ex); & \tikz\draw[red,fill=red] (0,0) circle (.7ex); & \tikz\draw[red,fill=red] (0,0) circle (.7ex); & \tikz\draw[red,fill=red] (0,0) circle (.7ex); & \tikz\draw[red,fill=red] (0,0) circle (.7ex); & \tikz\draw[blue,fill=blue] (0,0) circle (.7ex); \\

        2021 & Gui et al \cite{gui2021review} & \tikz\draw[green,fill=green] (0,0) circle (.7ex); & \tikz\draw[green,fill=green] (0,0) circle (.7ex); & \tikz\draw[green,fill=green] (0,0) circle (.7ex); & \tikz\draw[green,fill=green] (0,0) circle (.7ex); & \tikz\draw[blue,fill=blue] (0,0) circle (.7ex); & \tikz\draw[blue,fill=blue] (0,0) circle (.7ex); & \tikz\draw[red,fill=red] (0,0) circle (.7ex); & \tikz\draw[red,fill=red] (0,0) circle (.7ex); & \tikz\draw[red,fill=red] (0,0) circle (.7ex); \\
        
        2021 & Li et al \cite{li2021theoretical} & \tikz\draw[green,fill=green] (0,0) circle (.7ex); & \tikz\draw[green,fill=green] (0,0) circle (.7ex); & \tikz\draw[blue,fill=blue] (0,0) circle (.7ex); & \tikz\draw[blue,fill=blue] (0,0) circle (.7ex); & \tikz\draw[red,fill=red] (0,0) circle (.7ex); & \tikz\draw[red,fill=red] (0,0) circle (.7ex); & \tikz\draw[red,fill=red] (0,0) circle (.7ex); & \tikz\draw[red,fill=red] (0,0) circle (.7ex); & \tikz\draw[red,fill=red] (0,0) circle (.7ex); \\
        
        2022 & Xia et al. \cite{xia2022gan} & \tikz\draw[green,fill=green] (0,0) circle (.7ex); & \tikz\draw[green,fill=green] (0,0) circle (.7ex); & \tikz\draw[blue,fill=blue] (0,0) circle (.7ex); & \tikz\draw[red,fill=red] (0,0) circle (.7ex); & \tikz\draw[red,fill=red] (0,0) circle (.7ex); & \tikz\draw[red,fill=red] (0,0) circle (.7ex); & \tikz\draw[red,fill=red] (0,0) circle (.7ex); & \tikz\draw[red,fill=red] (0,0) circle (.7ex); & \tikz\draw[red,fill=red] (0,0) circle (.7ex); \\

        2022 & Xun et al. \cite{xun2022generative} & \tikz\draw[blue,fill=blue] (0,0) circle (.7ex); & \tikz\draw[red,fill=red] (0,0) circle (.7ex); & \tikz\draw[red,fill=red] (0,0) circle (.7ex); & \tikz\draw[red,fill=red] (0,0) circle (.7ex); & \tikz\draw[red,fill=red] (0,0) circle (.7ex); & \tikz\draw[green,fill=green] (0,0) circle (.7ex); & \tikz\draw[red,fill=red] (0,0) circle (.7ex); & \tikz\draw[red,fill=red] (0,0) circle (.7ex); & \tikz\draw[red,fill=red] (0,0) circle (.7ex);\\

        2023 & Ji et al. \cite{ji2023survey} & \tikz\draw[red,fill=red] (0,0) circle (.7ex); & \tikz\draw[green,fill=green] (0,0) circle (.7ex); & \tikz\draw[red,fill=red] (0,0) circle (.7ex); & \tikz\draw[red,fill=red] (0,0) circle (.7ex); & \tikz\draw[green,fill=green] (0,0) circle (.7ex); & \tikz\draw[red,fill=red] (0,0) circle (.7ex); & \tikz\draw[red,fill=red] (0,0) circle (.7ex); & \tikz\draw[red,fill=red] (0,0) circle (.7ex); & \tikz\draw[red,fill=red] (0,0) circle (.7ex); \\

        2023 & Iglesias et al. \cite{iglesias2023survey} & \tikz\draw[blue,fill=blue] (0,0) circle (.7ex); & \tikz\draw[green,fill=green] (0,0) circle (.7ex); & \tikz\draw[green,fill=green] (0,0) circle (.7ex); & \tikz\draw[blue,fill=blue] (0,0) circle (.7ex); & \tikz\draw[red,fill=red] (0,0) circle (.7ex); & \tikz\draw[blue,fill=blue] (0,0) circle (.7ex); & \tikz\draw[red,fill=red] (0,0) circle (.7ex); & \tikz\draw[red,fill=red] (0,0) circle (.7ex); & \tikz\draw[blue,fill=blue] (0,0) circle (.7ex); \\

        2023 & Brophy et al. \cite{brophy2023generative} & \tikz\draw[green,fill=green] (0,0) circle (.7ex); & \tikz\draw[green,fill=green] (0,0) circle (.7ex); & \tikz\draw[red,fill=red] (0,0) circle (.7ex); & \tikz\draw[red,fill=red] (0,0) circle (.7ex); & \tikz\draw[red,fill=red] (0,0) circle (.7ex); & \tikz\draw[red,fill=red] (0,0) circle (.7ex); & \tikz\draw[green,fill=green] (0,0) circle (.7ex); & \tikz\draw[red,fill=red] (0,0) circle (.7ex); & \tikz\draw[red,fill=red] (0,0) circle (.7ex);\\      \hline

        2023+ & Our survey & \tikz\draw[green,fill=green] (0,0) circle (.7ex); & \tikz\draw[green,fill=green] (0,0) circle (.7ex); & \tikz\draw[green,fill=green] (0,0) circle (.7ex); & \tikz\draw[green,fill=green] (0,0) circle (.7ex); & \tikz\draw[green,fill=green] (0,0) circle (.7ex); & \tikz\draw[green,fill=green] (0,0) circle (.7ex); & \tikz\draw[green,fill=green] (0,0) circle (.7ex); & \tikz\draw[green,fill=green] (0,0) circle (.7ex); & \tikz\draw[green,fill=green] (0,0) circle (.7ex); \\ \hline
    \end{tabular}
    \label{GAN_Survey}
\end{table*}

GANs have found applications beyond image generation, including video production and text generation. Vondrick et al. proposed a video generation GAN (VGAN) in 2018 \cite{vondrick2018tracking}, capable of producing realistic and diverse videos by learning to track and anticipate object motion. The VGAN architecture consisted of a motion estimation network and a video-generating network, jointly trained to generate high-quality videos. The results showcased VGAN's ability to produce realistic and varied films, enabling applications like video prediction and synthesis. Text generation is another domain where GAN has been utilized. In 2017, Yu et al. introduced SeqGAN, a GAN-based text generation model \cite{yu2017seqgan}. SeqGAN achieved realistic and diverse text generation capabilities by maximizing a reinforcement learning goal. The model included a generator responsible for text creation and a discriminator assessing the quality of the generated text. Through reinforcement learning, the generator was trained to maximize the predicted reward based on the discriminator's evaluation. The findings demonstrated that SeqGAN outperformed previous text generation algorithms, producing text that was more varied and lifelike. These advancements in GAN applications for video and text generation highlight the versatility and potential of GAN frameworks in diverse domains.

Another popular area of research focuses on addressing medical questions using GANs, as highlighted in the recent paper by Tan et al. where a GAN-based scale invariant post-processing approach is proposed for lung segmentation in CT Scans \cite{tan2021lgan}. A similar framework called RescueNet, developed by Nema et al., combines domain-specific segmentation methods and general-purpose adversarial learning for segmenting brain tumors \cite{nema2020rescuenet}. Their study not only suggests a promising technique for brain tumor segmentation but also advances the development of systems capable of answering complex medical inquiries. Despite the significant breakthroughs, there are still unresolved issues in GAN architectures and applications. One prominent challenge is the instability of GAN training, which can be influenced by various factors such as architecture, loss function, and optimization technique. In 2017, Arjovsky et al. proposed a solution called Wasserstein GAN (WGAN) \cite{arjovsky2017towards}, introducing a novel loss function and optimization algorithm to address stability issues in GAN training. Their approach showed improved stability and performance on datasets like CIFAR-10 \cite{abouelnaga2016cifar} and ImageNet \cite{recht2019imagenet}. 

\textbf{Related survey.} The existing body of research exploring various analytic tasks with GAN is accompanied by numerous surveys, which predominantly concentrate on specific perspectives within constrained domains, particularly computer vision and natural language processing. For instance, the survey by Jabbar et al. \cite{jabbar2021survey} explores applications of GANs in various industries, including computer vision, natural language processing, music, and medicine. To demonstrate the influence and promise of GANs in certain application domains, they also highlight noteworthy academic publications and real-world instances. The study tackles the difficulties and problems related to GAN training in addition to discussing their variations. The authors \cite{jabbar2021survey} investigate several training strategies, including minimax optimization, training stability, and assessment measures. They examine the typical challenges that arise during GANs training, such as mode collapse and training instability, and they give numerous solutions that have been suggested by researchers to address these problems. However, it does not specifically concentrate on GAN-based methods for imbalanced, time series, geoscience, and other data types and fails to reflect the most recent advancements in the field. The survey by Xia et al. \cite{xia2022gan} focuses on two primary categories of techniques for GAN inversion: Optimization-based methods and Reconstruction-based methods. To locate the hidden code that optimally reconstructs the supplied output, optimization-based approaches formulate an optimization issue. Reconstruction-based approaches, on the other hand, use different methods, such as feature matching or autoencoders, to directly estimate the latent code. An in-depth discussion of these strategies' advantages, disadvantages, and trade-offs is provided in the article. The non-convexity of the optimization issue and the lack of ground truth data for assessment are only two of the difficulties faced in GAN inversion that are highlighted in this article. The authors \cite{xia2022gan} additionally go through specific evaluation standards and measures designed for computer vision tasks. In addition, the study discusses current developments and variants in GAN inversion, such as techniques for managing conditional GAN, detaching latent variables, and dealing with different modalities. Aspect modification, domain adaptability, and unsupervised learning are a few of the applications and potential future directions of GAN inversion that are covered. A recent study by Durgadevi et al. \cite{durgadevi2021generative} presents a comprehensive overview of numerous GAN variants that have been proposed until 2020. Since its inception, GANs have undergone significant evolutions leading researchers to propose various enhancements and modifications aimed at addressing the prevalent challenges. These alterations encompass diverse aspects such as architectural design, training methods, or a combination of both. In this survey \cite{durgadevi2021generative} the authors delve into the application and impact of GANs in different domains including image processing, medicine, face detection, and text transferring. The survey by Alom et al. \cite{alom2019state} covers various aspects of the deep learning paradigm, such as fundamental ideas, algorithms, architectures, and contemporary developments including convolutional neural networks (CNNs), recurrent neural networks (RNNs), deep belief networks (DBNs), generative models, transfer learning, and reinforcement learning. The survey of Nandhini et al. \cite{nandhini2021deep} offers a thorough investigation of the application of deep CNNs and deep GANs in computational image analysis driven by visual perception. The designs and methodology used, the outcomes of the experiments, and possible uses for these approaches are covered in the paper. Overall this study provides a retrospective review of the development of GANs for the deep learning-based image analysis community. The survey by Kulkarni et al. \cite{kulkarni2019survey} presents an overview of various strategies, techniques, and developments used in GAN-based music generation. The survey of Sampath et al. \cite{sampath2021survey} summarizes the current advances in the GAN landscape for computer vision tasks including classification, object detection, and segmentation in the presence of an imbalanced dataset. Another survey by Brophy et al. \cite{brophy2023generative} attempts to review various discrete and continuous GAN models designed for time series-related applications. The study by Xun et al. \cite{xun2022generative} reviews more than 120 GAN-based models designed for region-specific medical image segmentation that were published until 2021. Another recent survey by Ji et al. \cite{ji2023survey} summarizes the task-oriented GAN architectures developed for symbolic music generation but other application domains are overlooked. The survey by Wang et al. \cite{wang2021generative} reviews various architecture-variant and loss-variant GAN frameworks designed for addressing practical challenges relevant to computer vision tasks. Another survey by Gui et al. \cite{gui2021review} provides a comprehensive review of task-oriented GAN applications and showcases the theoretical properties of GAN and its variants. The study by Iglesias et al. \cite{iglesias2023survey} summarizes the architecture of the latest GAN variants, optimization of the loss functions, and validation metrics in some promising application domains including computer vision, language generation, and data augmentation. The survey by Li et al. \cite{li2021theoretical} reviews the theoretical advancements in GAN and also provides an overview of the mathematical and statistical properties of GAN variants. A detailed comparison between our survey and others is presented in Table \ref{GAN_Survey}.

Although there are several papers reviewing GAN architecture and its domain-specific applications, none of them concurrently emphasize on applications of GAN in geoscience, urban planning, data privacy, imbalanced learning, and time series problems in a comprehensive manner. Methods developed to deal with these practical problems are underrepresented in past surveys. 
Moreover, the stability of GANs training, assessment of the produced data, and ethical issues with GAN are some of the issues that still need to be resolved. To fully exploit the future potential of GANs, more study in these areas is required. To fill the gap, this survey offers a comprehensive and up-to-date review of GANs, encompassing mainstream tasks ranging from audio, video, and image analysis, to natural language processing, privacy, geophysics, and many more. Specifically, we first provide several applied areas of GAN and discuss existing works from task and methodology-oriented perspectives. Then, we delve into multiple popular application sectors within the existing research of GAN with their limitations and propose several potential future research directions. Our survey is intended for general machine learning practitioners interested in exploring and keeping abreast of the latest advancements in GAN for multi-purpose use. It is also suitable for domain experts applying GANs to new applications or exploring novel possibilities building on recent advancements. 

\begin{figure*}
    \centering
    \includegraphics[scale=0.5]{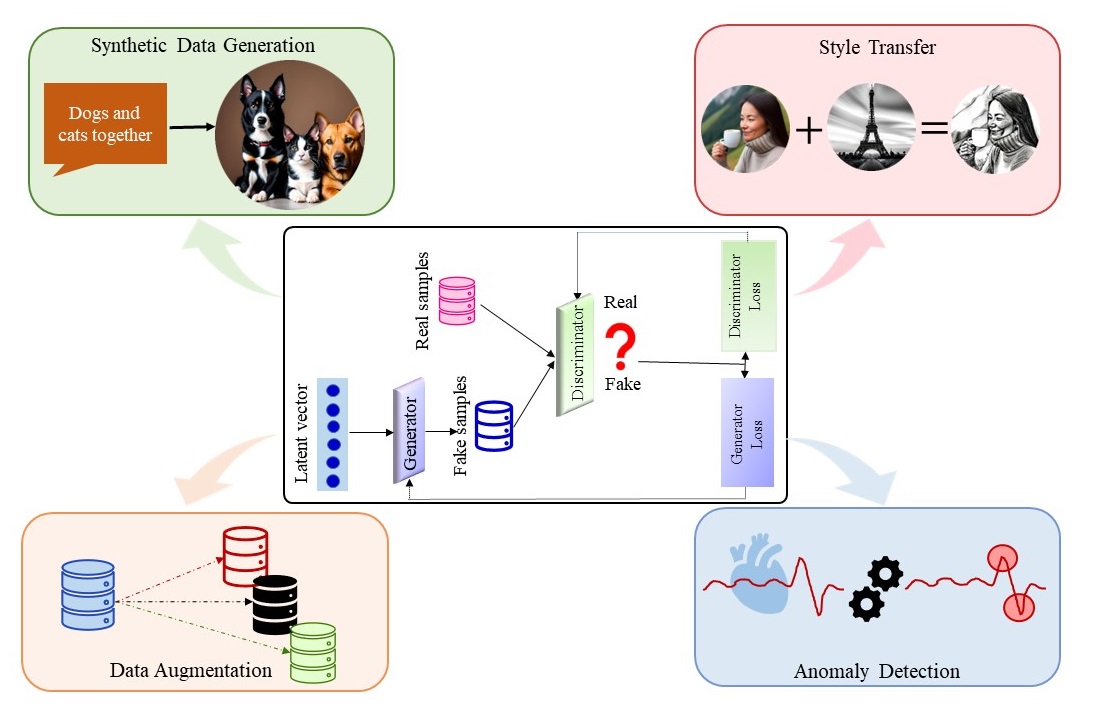}
    \caption{Architecture of GANs and its primary functions. In this example, different analytical tasks of GANs are categorized into synthetic data generation, style transfer, data augmentation, and anomaly detection.}
    \label{fig_no1}
\end{figure*}

\section{Overview of Generative Adversarial Network} \label{Section_Overview_GAN}

Generative Adversarial Networks (GANs) signify a pivotal advancement in artificial intelligence, offering a robust framework to craft synthetic data that closely resembles real-world information \cite{goodfellow2020generative}. Consisting of two interconnected neural networks, the Generator and Discriminator, GANs engage in a dynamic adversarial process that is redefining the landscape of deep generative modeling \cite{goodfellow2014generative, goodfellow2016deep}. By orchestrating this interplay, GANs transcend data generation frontiers across various domains, from crafting images to generating language, demonstrating a profound influence on reshaping the way machines comprehend and replicate intricate data distributions. This dynamic is facilitated through the Generator ($G$) network, entrusted with producing new data samples based on the input data distribution, while the Discriminator ($D$) network is devoted to discerning genuine data from their synthetic counterparts.

From a mathematical viewpoint, the $G$ network considers a latent space $z$ from the noise distribution $p_{z}$ as input and generates synthetic samples $G(z)$. Its goal is to generate data that is indistinguishable from real data samples $x$ originating from the probability distribution $p_{\text{data}}$. On the other hand, $D$ takes both real data samples $x$ from the actual dataset and fake data samples $G(z)$ generated by $G$ as input and classifies whether the input data is real or fake. It essentially acts as a ``critic'' that evaluates the quality of the generated data. The training process consists of both networks working in a two-player zero-sum game \cite{iglesias2023survey}. While $G$ aims to produce more realistic outcomes, $D$ enhances its ability to distinguish between real and fake samples. This dynamic prompts both players to evolve in tandem: if $G$ generates superior outputs, it becomes tougher for $D$ to discern them. Conversely, if $D$ becomes more accurate, $G$ faces greater difficulty in deceiving $D$. This process resembles a minimax game, where $D$ strives to maximize accuracy while $G$ seeks to minimize it \cite{goodfellow2016nips}. The goal is to find a balance where $G$ produces increasingly convincing data while $D$ becomes better at classifying real data from fake ones. The mathematical expression of this minimax loss function can be represented as:

\begin{equation}\label{GAN_loss}
    \underset{G}{\operatorname{min}} \; \underset{D}{\operatorname{max}} \; L = \mathbb{E}_{x\sim p_{\text{data}}} \left[\log D(x)\right] + \mathbb{E}_{z \sim p_{z}}\left[\log(1 - D(G(z)))\right],
\end{equation}
where the probability values $D(\mathbf{x})$ and $D(G(\mathbf{z}))$ represent the discriminator's outputs for real and fake samples, respectively. The first term in Eq. (\ref{GAN_loss}) encourages $D$ to correctly classify real data by maximizing $\log D(x)$, whereas the second term encourages $G$ to produce realistic data that $D$ classifies as real by minimizing $\log(1 - D(G(z)))$. In essence, $G$ aims to minimize the loss while $D$ aims to maximize it, leading to a continual back-and-forth training process. Throughout the training, the generator's performance improves as it learns to generate more realistic data, and the discriminator's performance improves as it becomes better at distinguishing real from fake data. Ideally, this competition results in a generator that produces data that is virtually indistinguishable from real data, as judged by the discriminator. A visual representation of the GAN's architectural details and its primary functions is presented in Fig. \ref{fig_no1}. 

\begin{figure*}
    \centering
    \includegraphics[width=0.9\textwidth]{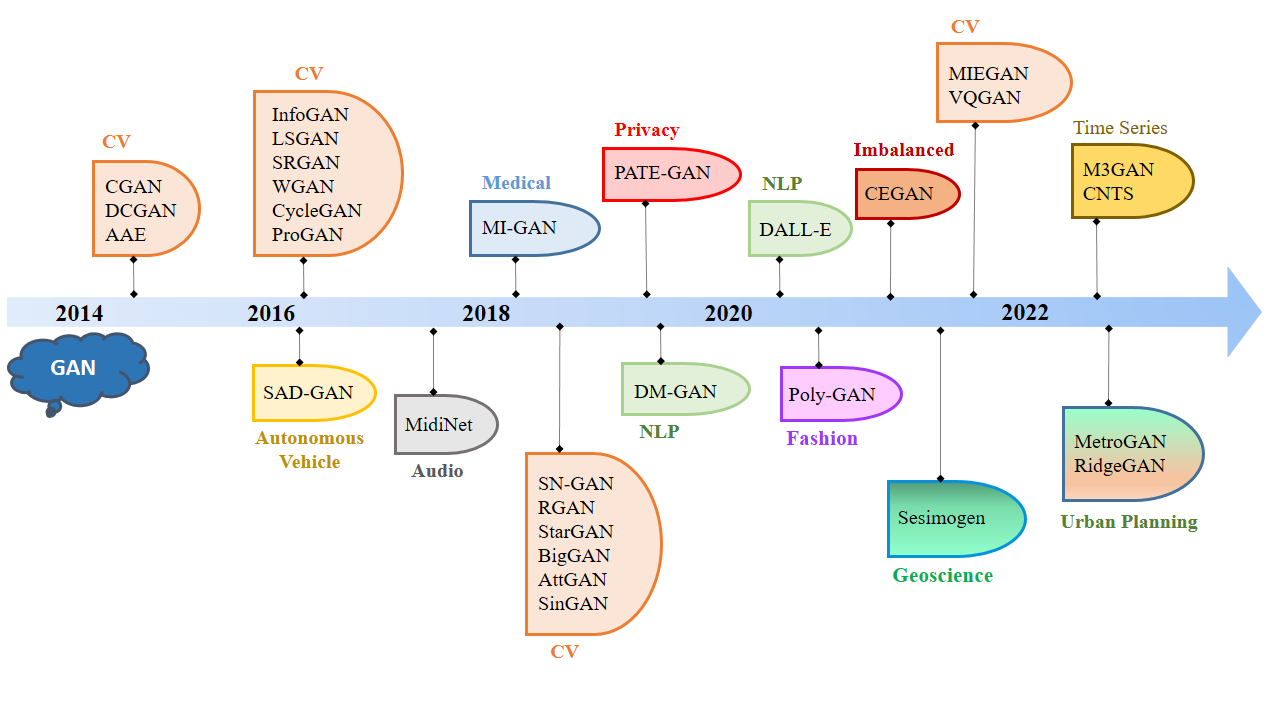}
    \caption{Timeline of the application-based GAN architectures reviewed in this study}
    \label{fig_no3}
\end{figure*}

During the time of the inception of GAN in 2014, Goodfellow et al. \cite{goodfellow2014generative} proved the existence of a unique solution for the minimax loss function. This solution became popular as Nash Equilibrium (NE) which reflects the equilibrium point where the generator's capacity to generate realistic data matches the discriminator's capacity to distinguish between real and fake data, resulting in high-quality synthetic data that closely resembles the true underlying data distribution \cite{nash1951non}. However, recent studies have revealed that attaining NE in GANs is not guaranteed and can be challenging due to various factors, including architecture choices, hyperparameters, and convergence difficulties \cite{heusel2017gans, farnia2020gans}. To address these challenges and enhance GAN's training stability researchers have developed various techniques, such as different loss functions and architectures over the decade \cite{liu2021generative}. These alterations of GAN include architectural changes, loss function-based modifications, and many others. They encompass various variations, each with unique attributes and applications, driving significant advancements in generative modeling. Fig. \ref{fig_no3} visually depicts the timeline of key developments in GAN research. 

\section{Application}\label{Section_Application}

\begin{figure}
    \centering
    \includegraphics[scale=0.45]{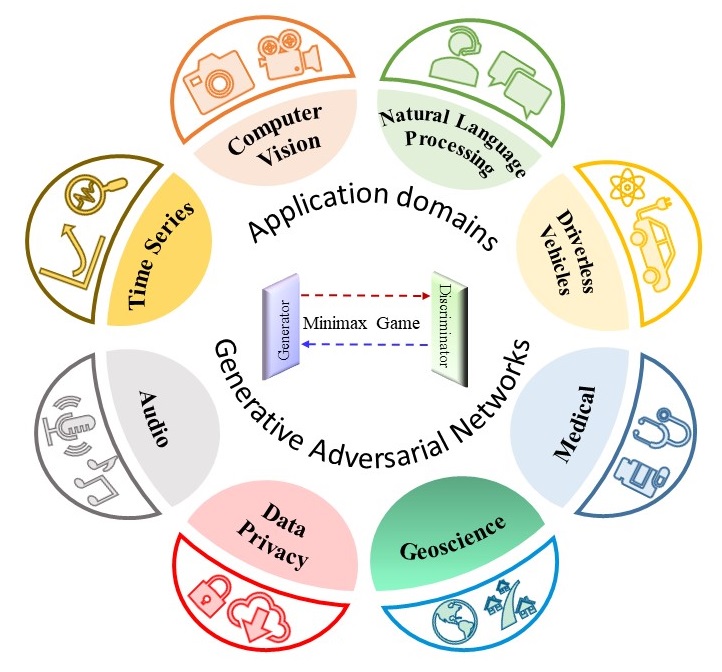}
    \caption{Diverse Applications of Generative Adversarial Networks (GANs) in various applied domains.}
    \label{fig_no2}
\end{figure}

As previously noted, GANs have emerged as one of the most prominent advancements in the realm of machine learning over recent years. GAN models have demonstrated their efficacy in domains where prior models fell short, while also substantially enhancing performance in other scenarios. Within this section, we will comprehensively explore the pivotal domains where GAN architectures have been deployed. While much of the recent research has concentrated on employing GANs to generate novel synthesized data, emulating distinct data distributions, our exploration in this section will highlight the broader applications of GANs, extending to areas such as video game development \cite{kim2020learning}, urban planning \cite{thottolil2023prediction} and others. We also visually showcase the application domains of GAN in Fig. \ref{fig_no2}.

\paragraph{Image Generation} Among the most promising domains harnessing the capabilities of GANs is computer vision. Notably, the generation of realistic images stands as one of the paramount applications of GANs \cite{karras2019style, cao2018recent}. The capacity of GANs to craft authentic images depicting characters, animals, and objects that lack real-world existence holds immense significance \cite{ma2017pose}. This capability of GAN finds application in diverse projects, spanning from refining facial recognition algorithms to fabricating immersive virtual environments for video games and commercial campaigns \cite{yu2017unsupervised}. Moreover, GANs have proven instrumental in generating true-to-life virtual realms, a boon for both the gaming industry and advertising ventures. By crafting synthetic landscapes and structures, GANs empower game designers and developers to construct captivating, realistic virtual worlds, thereby elevating the overall player experience \cite{karras2017progressive}. The deployment of GANs in this context offers a swift, cost-effective, and efficient alternative to traditional manual design and modeling approaches, enabling the production of high-quality graphics.

\paragraph{Video Synthesis} In addition to generating high-quality images, GANs offer the potential to create synthetic videos, a more complex task due to coherence requirements \cite{wang2020imaginator}. GANs, combining generators and discriminators, excel in this challenge \cite{tulyakov2018mocogan}. The discriminator learns to differentiate real from synthetic frames, while the generator produces visually authentic video frames. GANs find widespread use in replicating real-world actions, enhancing surveillance and animations \cite{wang2023videofactory}. One of the most popular and controversial applications of GAN is the evolution of Deepfake \cite{westerlund2019emergence}. Deepfakes are AI-generated media, that blend a person's likeness with another's context using GANs. While they offer creative potential, deepfakes raise ethical concerns, requiring a holistic approach to detect them \cite{korshunov2019vulnerability, yu2021survey}. 

\paragraph{Augmenting data} GANs possess the capability to generate synthetic data, which can be harnessed to bolster actual data and enhance the performance of deep learning models. This approach is instrumental in mitigating concerns related to data scarcity and refining model accuracy \cite{xie2020unsupervised}. GANs provide an effective avenue for fortifying machine learning and deep learning frameworks with authentic data. Addressing the challenge of limited data availability, GANs enable the creation of larger, more diverse datasets by generating artificial samples that closely emulate real data \cite{bowman2015generating}. GAN-based data augmentation strategies have showcased promising outcomes across various domains, offering the potential to enhance model precision and transcend the constraints posed by insufficient data \cite{frid2018synthetic}.

\paragraph{Style Transfer} GANs are capable of transferring the style of one image to another, resulting in the creation of an entirely new image \cite{johnson2016perceptual}. This method can be applied to develop novel artistic features or enhance the visual attractiveness of pictures. By facilitating the development of fresh artistic trends and boosting the aesthetic appeal of pictures, GAN-based style transfer approaches have transformed the area of computer vision \cite{zhu2017unpaired,gatys2015neural}. These methods have been used in a variety of fields, such as digital art, photography, and graphic design, and they continue to be an inspiration for new developments and studies in the area.

\paragraph{Natural Language Processing} Over the past few years, GANs have been adapted to process text data, resulting in groundbreaking advancements within the realm of Natural Language Processing (NLP). One notable application involves text generation, where GANs can create coherent and contextually relevant textual content. For instance, the Text GAN framework utilizes Long Short-Term Memory (LSTM) networks \cite{hochreiter1997long} as the generator and CNN as the discriminator to synthesize novel text using adversarial training \cite{zhang2016generating}. Furthermore, GANs play a role in text style transfer, allowing alterations in writing styles while preserving content, and enhancing the adaptability of generated material \cite{toshevska2021review}. In the domain of sentiment analysis, GANs contribute by generating text with specific emotional tones, thereby aiding model training and dataset augmentation for sentiment classification tasks. Additionally, GANs are instrumental in text-to-image synthesis, translating textual descriptions into visual representations, proving valuable in fields like accessibility and multimedia content creation \cite{zhang2017stackgan}. GANs have also been harnessed to enhance machine translation software, refining translation precision and fluidity \cite{yu2017seqgan,guo2018long}.

\paragraph{Music Generation} GANs are revolutionizing music creation by tapping into existing compositions' patterns and structures \cite{mu2021review}. This technology not only fosters original music composition but also assists musicians in their creative journey. Previous studies have showcased GANs' role in generating music, offering possibilities for both novel compositions and artist support \cite{dong2018musegan, civit2022systematic}. Beyond composition, GANs empower musicians to explore new styles by generating melodies, harmonies, and rhythms as creative sparks. They also enable style transfer, allowing musicians to reimagine their music in diverse genres and cultural contexts. Moreover, GANs have ventured into musical collaboration, aiding improvisation by responding to musician input with harmonious suggestions. In essence, GANs redefine music creation, from assisting composers in originality to fostering innovative style exploration \cite{mao2018semantic}. This fusion of human creativity and computational ability promises to shape the future of the music industry.

\paragraph{Medical Domain} In the dynamic landscape of the medical domain, GANs have emerged as a game-changing technology with multifaceted benefits. The integration of GANs with medical data holds immense potential in enhancing disease diagnosis through the creation of synthetic medical images thereby eliminating the limited data problem. This expanding diversity and quantity of data made possible by GANs empower the data-driven diagnostic models to deliver more precise and reliable predictions, aiding healthcare practitioners in making accurate diagnoses and ultimately enhancing patient care \cite{guibas2017synthetic, singh2021medical, wang2021dicyc}. Another significant application of GAN is in drug discovery, where it can process and generate molecular structures with desired properties \cite{kadurin2017cornucopia, kadurin2017drugan}. GAN-driven molecular generation accelerates the process of identifying potential drug candidates, saving time and resources in the search for novel therapeutic compounds. Moreover, GANs extend their impact to surgical training and planning by producing realistic surgical scenarios and simulations \cite{zhao2022surgical} and also aid in generating patient-specific medical images, allowing healthcare practitioners to tailor treatment plans to individual patient characteristics \cite{ma2020feasibility}.
 
\paragraph{Urban Planning} With rapid urbanization, predicting transportation patterns is essential for sustainable urban planning and traffic management. Recent advancements in GAN-based methods to simulate hyper-realistic urban patterns, including CityGAN \cite{Albert2018ModelingUP}, Conditional GAN with physical constraints \cite{albert2019spatial}, and MetroGAN \cite{zhang2022metrogan}, have become popular in urban science fields. These GANs can generate synthetic urban universes that mimic global urban patterns, and quantifying landscape structures of these GAN-generated new cities using spatial pattern analysis helps in understanding landscape dynamics and improving sustainable urban planning. In a recent study, a novel RidgeGAN model \cite{thottolil2023prediction} is proposed that evaluates the sustainability of urban sprawl associated with infrastructure development and transportation systems in medium and small-sized cities.

\paragraph{Geoscience and Remote Sensing} In geoscience, there are also recent applications of GANs with novel ways of generating ``new'' samples that can easily outperform state-of-the-art geostatistical tools. This is very appealing in applications like reservoir modeling as geologists and reservoir engineers are nowadays usually tasked to work with multiple realizations of the subsurface and provide probabilistic estimates to support the subsequent decision-making process. A few examples of early applications of GANs in geoscience are the reconstruction of three-dimensional porous media \cite{mosser2017reconstruction}; Generating geologically realistic 3D reservoir facies models using deep learning of sedimentary architecture \cite{zhang2019generating}; and SeismoGen: Seismic Waveform Synthesis Using GAN With Application to Seismic Data Augmentation \cite{wang2021seismogen}.

\paragraph{Autonomous Vehicles} Machine learning models for autonomous driving can be trained using synthetic pictures of real-world situations created using GANs. This method helps to mitigate the safety concerns of autonomous cars by getting beyond the restrictions of real-world testing \cite{gecer2018semi}. A potential method for training autonomous driving models is the use of GANs to produce synthetic visuals \cite{pan2017virtual}. It makes it possible to investigate a wide range of complex scenarios, improving the performance and safety of the models. Recent studies have illustrated the usefulness and promise of this method for bridging the gap between driving simulations and actual driving situations, ultimately promoting the development of autonomous cars \cite{shrivastava2017learning, zhang2018deeproad}.

\paragraph{Fashion and design} GANs find utility in generating fresh patterns and designs for clothing, aiding designers in crafting innovative collections. This technology extends its impact on online shopping experiences by producing images of apparel on virtual models, offering customers a realistic preview of how garments would appear on them during online purchases \cite{jiang2017fashion}. Within the realms of fashion and design, GANs have become a valuable asset, empowering designers to stretch their creative boundaries by facilitating the creation of novel patterns and designs \cite{han2018viton}. Furthermore, GAN-driven virtual try-on systems enhance the convenience of online shopping, granting shoppers lifelike insights into how clothing would fit and appear on them. Several diverse research efforts in this domain have explored the significant contributions of GAN in the evolution of the fashion and design industry \cite{liu2019toward, pandey2020poly}.

\paragraph{Imbalanced Pattern Classification} A prevalent yet intricate issue encountered in pattern recognition is referred to as ``class imbalance'', signifying disparities in the frequencies of class labels \cite{chakraborty2020hellinger}. To address this challenge, GANs can be used to generate synthetic data for the minority class of various imbalanced datasets as a method of intelligent oversampling \cite{dam2022latent}. Pioneering approaches such as Balancing GAN (BAGAN) \cite{mariani2018bagan} and Classification Enhancement GAN (CEGAN) \cite{suh2021cegan} have been developed to restore balance in the distributions of imbalanced datasets and enhance the precision of the data-driven models.

\paragraph{Time Series Anomaly Detection} In recent years there has been a significant surge in the availability of real-time sensor data across diverse domains including healthcare systems, power plants, industries, and many others. These vast datasets are often accompanied by several anomalous events which eventually diminishes the modeling capabilities of any machine learning and deep learning frameworks. To address this issue anomaly detection for multivariate time series data has become a critical task for time series analysts \cite{panja2023epicasting}. In this context, GANs have become a powerful technology. In recent studies, various GAN-based time series anomaly detection techniques namely, Dilated Convolutional Transformer GAN (DCT GAN) \cite{li2021dct}, M2GAN \cite{li2023m3gan}, Cooperative Network Time Series (CNTS) \cite{yang2023cnts}, TADGAN \cite{geiger2020tadgan}, and many others have been developed that leverage the power of adversarial training to efficiently detect the presence of anomalous data. 

\paragraph{Data privacy} GANs offer the possibility of generating synthetic data that retains the statistical characteristics of the original data, all while safeguarding sensitive information. This approach serves as a means to ensure privacy protection for individuals while enabling the secure utilization of data for research and analytical purposes \cite{liu2019ppgan}. A recent study by Torfi et al. has demonstrated how GAN can be leveraged to generate synthetic data that mimics the statistical properties of the real dataset thus preserving data privacy \cite{torfi2020corgan}. This development creates new opportunities for private data sharing and analysis, offering insightful information while preserving privacy.

In conclusion, GANs have a wide range of applications across diverse domains, from generating realistic images and movies to aiding in medical diagnosis \cite{goodfellow2014generative, karras2019style}. The restrictions of data scarcity can be eliminated, and personal information can be safeguarded, by developing synthetic data that closely resembles actual data \cite{shokri2017membership}. As GANs develop further, we can witness more cutting-edge applications in real-data problems \cite{radford2015unsupervised}. In summary, GANs offer a wide range of applications in a variety of sectors and have the ability to completely change how we produce and use data \cite{gatys2016image, arjovsky2017wasserstein}. Future GAN applications are likely to have even more fascinating uses as the technology develops \cite{brock2018large}.

\section{Variants OF GAN} \label{gan_vARIANTS_SECTION}
In this section, we will have a broad review of some of the GAN models based on their distinct characteristics and practical uses. Additionally, we discuss the mathematical formulation of these GAN variants, using standard notations as discussed in Sec. \ref{Section_Overview_GAN} and present their implementation software in Table \ref{tab:software-libraries}. \\
  
\textbf{CGAN.}
The conditional GAN (CGAN) is a popular version of GAN that generates data by taking external inputs, such as labels or classes, into account. It was introduced by Mirza and Osindero in 2014 \cite{mirza2014conditional} and has since been widely used in computer vision applications, including image synthesis, image-to-image translation, and text-to-image synthesis. Unlike the conventional GAN both $G$ and $D$ of the CGAN architecture receive conditional information $y$ that serves as a guide for $G$ to produce data that aligns with the specified conditions.  The loss function for the CGAN framework is given by:
$$
L = \mathbb{E}_{x\sim p_{\text{data}}} \left[\log D(x, y)\right] + \mathbb{E}_{z \sim p_{z}}\left[\log(1 - D(G(z, y), y))\right]. $$
The CGAN model, as discussed in the literature \cite{mirza2014conditional, odena2017conditional}, possesses the following key features:
\begin{itemize}
    \item CGANs generate customized data that is specific to a given input, e.g., a CGAN trained on animal photos can produce images of a particular animal based on the input.
    \item Unlike Vanilla GAN, CGAN benefits from additional inputs, resulting in synthetic data of higher quality. It exhibits improved coherence, structure, and aesthetic resemblance to real samples.
    \item CGANs demonstrate superior noise resistance compared to other artificial neural networks due to the utilization of external input to guide the data generation process.
\end{itemize}
While the CGAN model is known for its versatility, it is also accompanied by several limitations. It is prone to overfitting with scarce or noisy input data, requires explicit labels or classes in the input dataset, is vulnerable to adversarial attacks, and becomes computationally complex with high-dimensional complex datasets \cite{szegedy2013intriguing}. Considering both the advantages and disadvantages of the CGAN model mentioned above, it proves to be a valuable tool for generating data based on external input \cite{xiao2022generative}. However, it is important to take into account these limitations and drawbacks when applying CGANs to address specific problems. Future research can examine alternative conditioning methods including the use of natural language descriptions or a variety of circumstances \cite{denton2015deep}.\\

\textbf{DCGAN.}
Deep Convolutional GAN (DCGAN) introduced by Radford et al. in 2015 \cite{radford2015unsupervised} marks a significant breakthrough in the realm of generative AI, particularly for image generation. Representing a specialized variation of the GAN architecture, DCGANs seamlessly combine CNN and GAN techniques to yield high-quality, photorealistic images with intricate details. With the ability to autonomously learn and generate images without additional control, DCGANs prove their usefulness in unsupervised learning scenarios. DCGANs stand out for their relatively manageable training process, owing to sophisticated architectural components like strided convolutions, batch normalization, and leaky Rectified Linear Unit (ReLU) activation functions \cite{radford2015unsupervised}. From the experimental perspective, DCGANs have generated excellent results for large-scale picture datasets like CIFAR-10 and ImageNet,  \cite{krizhevsky2009learning}. Nonetheless, it is worth noting that DCGANs exhibit elevated computational demands, sensitivity to hyperparameters, and susceptibility to challenges such as restricted diversity of generated images and mode collapse \cite{lucic2018gans}. Despite these limitations, DCGANs find successful applications across domains encompassing image synthesis, style transfer, and image super-resolution. Their far-reaching impact on the field of generative modeling continues to inspire advancements and innovation.\\

\textbf{AAEs.} Adversarial Autoencoder (AAE) framework, proposed by Makhzani et al. in 2015, is a hybridization of autoencoders with adversarial training \cite{makhzani2015adversarial}. This model has garnered significant attention due to its potential for variational inference by aligning the aggregated posterior of the hidden code vector with a chosen prior distribution. This approach ensures that meaningful outcomes emerge from various regions of the prior space. Consequently, the AAE's decoder acquires the capability to learn a sophisticated generative model, effectively mapping the imposed prior to the data distribution. AAEs excel in producing disentangled representations, showcasing noise resistance, and generating high-quality images. The components within the AAE framework offer notable advantages over alternative generative models. Through adversarial training, AAEs excel in capturing complex data distributions and generating detailed, high-quality images. Their ability to learn disentangled representations in separate latent dimensions empowers precise image control, encompassing alterations to object properties. AAEs exhibit resilience to input variations, making them valuable for noisy data scenarios. Their encoder-decoder design supports denoising and surpasses other models in semi-supervised classification \cite{makhzani2015adversarial}. However, like other generative models, AAEs can encounter mode collapse, demand substantial computational resources, and necessitate cautious hyperparameter tuning. Striking the right balance between adversarial training and autoencoder loss poses a challenge. AAEs lack explicit control over generated samples, hindering targeted data traits in fine-grained control contexts \cite{bousmalis2017unsupervised}. Yet, the application scope of AAEs is notably expanded by the enhanced encoder, decoder, and discriminator networks, even surpassing traditional autoencoders.\\

\textbf{InfoGAN.}
Information Maximizing Generative Adversarial Network (InfoGAN), a modification of GAN, is designed to learn disentangled representations of data by maximizing the mutual information between a subset of the generator's input and the generated output. It was introduced by Chen et al. in 2016 \cite{chen2016infogan}. The loss function formulation for the Generator in InfoGAN is as follows:
\begin{multline*}
    L = \mathbb{E}_{x\sim p_{\text{data}}} \left[\log D(x)\right] + \mathbb{E}_{z \sim p_{z}}\left[\log(1 - D(G(z)))\right] \\ - \lambda\mathcal{I}(c;G(z)),\end{multline*}
where $\mathcal{I}(c; G(z))$ is the mutual information between the generator's output $G(z)$ and the learned latent code $c$, and $\lambda$ is a hyperparameter that regulates the trade-off between the adversarial loss and the mutual information term. 
The information-theoretic approach employed in the InfoGAN framework enhances its ability to learn representations that facilitate data exploration, interpretation, and manipulation tasks. Unlike supervised methods, InfoGAN does not rely on explicit supervision or labeling, making it a flexible and scalable option for unsupervised learning tasks like image generation and data augmentation. However, the InfoGAN framework may struggle to learn meaningful and interpretable representations for high-dimensional complex datasets, and its benefits may not always justify the additional complexity and computational cost. Overall, InfoGAN shows promising results in learning disentangled representations, but its effectiveness depends on specific goals, data characteristics, and available resources \cite{higgins2017beta}. Ongoing research and advancements hold the potential to address limitations and further improve this approach in the future.\\

\textbf{SAD-GAN.}
The Synthetic Autonomous Driving using GANs (SAD-GAN) model, introduced by Ghosh et al. in 2016, is designed to generate synthetic driving scenes using the GAN approach \cite{ghosh2016sad}. This model's core concept involves training a controller trainer network using images and keypress data to replicate human learning. To create synthetic driving scenes, the SAD-GAN is trained on labeled data from a racing game, consisting of images portraying a driver's bike and its surroundings. A key press logger software is employed to capture key press data during bike rides. The framework's architecture is inspired by DCGAN \cite{radford2015unsupervised}. The generator takes a current-time input image and produces the subsequent-time synthetic image. Meanwhile, the discriminator receives the real latest-time image, generates its feature map via convolution, and compares real and synthetic scenes to train the generator through a minimax game. The SAD-GAN framework offers an autonomous driving prediction algorithm suitable for manual driving as a recommendation system. Nevertheless, like DCGAN, it requires substantial computation and is susceptible to mode collapse, limiting its real-time applications.\\

\textbf{LSGAN.}
Traditional GAN models typically utilize a discriminator modeled as a classifier with the sigmoid cross entropy loss function. However, this choice of loss function can result in the issue of vanishing gradients during training, resulting in impaired learning of the deep representations. To address this concern, Mao et al. introduced a novel approach called Least Squares GAN (LSGAN) in 2017, which employs the least squares loss function for the discriminator instead \cite{mao2017least}. Mathematically, the Generator loss function $\left(L_G\right)$ and the Discriminator loss function $\left(L_D\right)$ of LSGAN model is expressed as follows:
\begin{align*}
    L_G &= \frac{1}{2}\mathbb{E}_{z\sim p_z}\left[(D(G(z)) - c)^2\right],   \\
    L_D &= \frac{1}{2} \mathbb{E}_{x \sim p_{\text{{data}}}}(D(x) - b)^2 + \frac{1}{2} \mathbb{E}_{z \sim p_{\mathbf{z}}}(D(G(z))-a)^2,
\end{align*}
where $a$-$b$ encoding scheme represents the labels for fake data and real data for $D$, and $c$ denotes the values that $G$ wants $D$ to believe for fake data. The LSGAN framework represents a notable advancement over traditional GANs, offering improved stability and convergence during training while generating higher-quality synthetic data. It has outperformed regular GANs in generating realistic images, as measured by Inception score, across various datasets such as CIFAR-10 \cite{mao2017least}. However, LSGANs often produce fuzzy images due to the use of squared loss in the objective function. The generated images often lack sharpness and fine details, as the loss function penalizes large discrepancies between fake and real images but neglects smaller variations. Researchers have addressed this issue by modifying the loss function in subsequent studies, aiming to enhance the sharpness of synthetic images \cite{gulrajani2017improved, huang2019ccnet}. While LSGANs show promise in generating high-quality images, ongoing research and development are focused on overcoming their limitations in producing crisp and detailed results.\\

\textbf{SRGAN.}
Super Resolution GAN (SRGAN), introduced by Ledig et al. in 2017, is a GAN-based framework for image super-resolution \cite{ledig2017photo}. It generates high-resolution images from low-resolution inputs with an upscaling factor of 4 using a generator network and a discriminator network. To achieve super-resolution, SRGAN incorporates a perceptual loss function, combining content and adversarial losses. Mathematically, the perceptual loss is expressed as:
$$
    l^{\text{SR}} = l_x^{\text{SR}} + 10^{-3} l_{\text{Gen}}^{\text{SR}}, 
$$
where $l_x^{\text{SR}}$ represents the content loss and $l_{\text{Gen}}^{\text{SR}}$ is the adversarial loss. The content loss used in the SRGAN framework relies on a pre-trained VGG-19 model and it provides the network information regarding the quality and content of the generated image. On the other hand, the adversarial loss is responsible for ensuring the generation of realistic images from the generator network.
SRGANs offer the ability to generate high-quality images with enhanced details and textures, resulting in improved overall image quality. They excel in producing visually appealing and realistic images, as confirmed by studies on perceptual quality \cite{johnson2016perceptual}. SRGANs exhibit noise resistance, enabling them to handle low-quality or noisy input images while still delivering high-quality outputs \cite{wang2004image}. Moreover, this model demonstrates flexibility and applicability across various domains, including video processing, medical imaging, and satellite imaging \cite{ledig2017photo}. However, training SRGANs can be computationally expensive, especially for complex models or large datasets. Additionally, like other GANs, the interpretability of SRGANs can be challenging, making it difficult to understand the underlying learning process of the generator. Furthermore, while SRGANs excel in image synthesis, they may not perform as effectively with text or audio inputs, limiting their range of applications.\\

\textbf{WGAN.}
The Wasserstein GAN (WGAN), introduced by Arjovsky et al. 
in 2017, is a loss function optimization variant of GAN that improves training stability and mitigates mode collapse \cite{arjovsky2017towards}. It employs the Wasserstein distance to enhance realistic sample generation and ensure meaningful gradients. By introducing a critic network and weight clipping, WGAN achieves training stability. It finds applications in image synthesis, style transfer, and data generation. The formulation of the WGAN framework utilizes the Wasserstein-1 distance or the Earth Mover distance to measure the distance between real and generated data distributions. Mathematically, the Wasserstein distance for transforming the distribution $\mathbb{P}$ to distribution $\mathbb{Q}$ can be expressed as:
$$
W(\mathbb{P}, \mathbb{Q}) = \underset{\theta \in \pi(\mathbb{P}, \mathbb{Q})} {\operatorname{inf}}\mathbb{E}_{(\tilde{X}, \tilde{Y}) \sim \theta} \left[\|\tilde{X} - \tilde{Y}\|\right].
$$
In the WGAN model, the discriminator function $D$ is designed as a critic network that estimates the Wasserstein distance between the real data distribution and the generated data distribution instead of probability values as in conventional GAN. These scores reflect the degree of similarity or dissimilarity between the input sample and the real data distribution. The training of the critic in WGAN involves optimizing its parameters to maximize the difference in critic values between real and generated samples. By clipping the discriminator weights, the discriminator loss function in WGAN is adjusted to enforce the Lipschitz continuity requirement, but the fundamental structure of the loss functions is maintained. In general, WGANs have demonstrated improved training stability compared to traditional GANs. They are less sensitive to hyperparameters and more resistant to mode collapse \cite{gulrajani2017improved}. The use of the Wasserstein distance facilitates smoother optimization and better gradient flow, resulting in faster training and higher-quality samples. However, calculating the Wasserstein distance can be computationally expensive \cite{mescheder2017numerics}. Although WGANs offer enhanced stability, careful tuning of hyperparameters and network designs is still necessary for satisfactory results. Furthermore, WGANs are primarily suited for generating images and may have limited applicability to other types of data. In summary, WGANs represent a promising advancement in the field of GANs, addressing their limitations and providing insights into distribution distances, but the applicability of WGANs to real-world problems requires careful consideration of its challenges.\\

\textbf{CycleGAN.}
Cycle-Consistent GAN (CycleGAN), introduced by Zhu et al. in 2017, is an unsupervised image-to-image translation framework that eliminates the need for paired training data unlike traditional GANs \cite{zhu2017unpaired}. It relies on cycle consistency, allowing images to be translated between two domains using two generators and two discriminators while preserving coherence. One generator $G_{XY}$ translates images from the source domain $X$ to the target domain $Y$, and the other $G_{YX}$ performs the reverse. In other words the function $G_{YX}$ is such that $G_{YX}(G_{XY}(x)) = x$. The discriminators, on the other hand, distinguish between real and translated images generated by the generators. To train this architecture the cycle consistency loss of Cycle GAN plays a crucial role by enforcing consistency between the original and round-trip translated images, the so-called \textit{forward} and \textit{backward} consistency. This ensures generators produce meaningful translations, preserving important content and characteristics across domains. Mathematically, the cycle consistency loss function can be expressed as:
\begin{multline*}
    \mathcal{L}_{\text{cycle}}(G_{XY}, G_{YX}) = \mathbb{E}_{x \sim p_{data}}[\|G_{YX}(G_{XY}(x)) - x\|_1] \\ + \mathbb{E}_{y \sim p_{data}}[\|G_{XY}(G_{YX}(y)) - y\|_1].
\end{multline*}
The main advantage of Cycle GAN lies in its ability to produce high-quality images with remarkable visual fidelity. It excels in various image-to-image translation tasks, including style transfer, colorization, and object transformation. Moreover, its computational efficiency allows training on large datasets. However, CycleGAN often suffers from mode collapse and the increasing amount of parameters reduces its efficiency \cite{sergio2016learning}. Despite its limitations, CycleGAN remains a valuable tool for image translation, and ongoing research for any data translation task aims to address its shortcomings \cite{yi2017dualgan}. For example, it shows promising results in medical imaging domain adaptation \cite{hashemi2018asymmetric}. \\

\textbf{ProGAN.}
In 2017, Karras et al. introduced the Progressive Growing of GAN (ProGAN), addressing the limitations of traditional GANs such as training instability and low-resolution output \cite{karras2017progressive}. ProGAN utilizes a progressive growth technique, gradually increasing the size and complexity of the generator and discriminator networks during training. This incremental approach enables the model to learn coarse characteristics first and subsequently refine them, ultimately producing high-resolution images. By starting with low-resolution image generation and progressively adding layers and details, ProGAN achieves training stability and generates visually realistic images of superior quality. This technique has found successful applications in various domains, including image synthesis, super-resolution, and style transfer. 
During training, the resolution of the generated images is increased progressively from a low resolution (e.g., 4x4) to a high resolution (e.g., 1024x1024). At each resolution level, the generator and discriminator networks are updated using a combination of loss functions. 
Progressive updates at increasing resolutions ensure high-quality image synthesis with fine features and textures throughout training, unlike the conventional GAN framework. ProGAN offers better scalability, enabling the generation of images at any resolution. It exhibits improved stability during training, overcoming issues like mode collapse. The flexibility of ProGAN makes it suitable for various image synthesis applications, including satellite imaging, video processing, and medical imaging \cite{karras2017progressive}. However, training ProGAN can be computationally expensive, especially for large datasets or complex models. Interpretability may pose challenges, as with other GANs, making it difficult to discern the learned representations. Additionally, ProGAN's generalization to new or unexplored data may be limited, requiring further fine-tuning or training on fresh datasets \cite{zhang2018unreasonable}.\\

\textbf{MidiNet.}
MidiNet, proposed by Yang et al. in 2017, attempts to generate melodies or a series of MIDI notes in the symbolic domain \cite{yang2017midinet}. Unlike other music generation frameworks, such as WaveNet \cite{oord2016wavenet}, and Song from PI \cite{chu2016song}, the MidiNet model can generate melodies either from scratch or by combining the melodies of previous bars. The architectural configuration of the MidiNet framework is motivated by the DCGAN model \cite{radford2015unsupervised}. The MidiNet model combines a CNN generator with a conditioner CNN in the first phase of training. While the former CNN is employed to generate synthetic melodies based on the random noise vector, the latter provides the available prior knowledge about other melodies in the form of an encoded vector as an optional input to the generator. Once the melody is generated it is processed with a CNN-based discriminator which consists of a few convolutional layers and a fully connected network. The discriminator is optimized using a cross-entropy loss function to efficiently detect whether the input is a real or a generated one. For training the overall network in MidiNet, the minimax loss function is combined with feature mapping and one-sided label smoothing to ensure learning stability and versatility in the generated content. The MidiNet framework proposes a unique CNN-GAN structure for the generation of symbolic melodies. Its ability to synthesize artificial music in the presence or absence of prior knowledge is very useful in the audio domain. However, due to the use of a CNN-based structure, its computational complexity significantly increases in comparison to the standard GAN model. Further research in this domain is required to understand the capabilities of MidiNet in multi-track music generation while simultaneously reducing its running time.\\

\textbf{SN-GAN.}
Spectral Normalization GAN (SN-GAN) is a GAN variant that utilizes spectral normalization to stabilize the training of the generator and discriminator networks \cite{miyato2018spectral}. In conventional GANs, training can be unstable due to a powerful discriminator or poor-quality generator samples. SN-GAN addresses this by constraining the Lipschitz constant of the discriminator, preventing it from dominating the training process. Spectral normalization normalizes the discriminator's weight matrices, ensuring a stable maximum value and preventing the amplification of minor input perturbations. SN-GAN produces high-quality samples with improved stability and convergence compared to traditional GANs.
The adversarial training process used in the SN-GAN framework, similar to the conventional GAN (as in Eq. \ref{GAN_loss}), encourages $G$ to produce more realistic samples that can fool $D$, while $D$ learns to accurately distinguish between real and generated samples. Several benefits of the SN-GAN model over the standard GAN include increased stability in training the generator and discriminator by constraining the Lipschitz constant of the discriminator. This mitigates issues like gradient explosion and mode collapse, resulting in high-quality examples with fine features and edges. SN-GAN is relatively simple to implement and can be integrated into existing GAN systems. However, the computation of singular values during the normalization process adds to the computational burden, potentially extending training time and requiring more memory. SN-GAN's reliance on the spectral norm assumption of discriminator weights may limit its applicability to specific GAN architectures. While SN-GANs may exhibit slower convergence and reduced sample diversity compared to conventional GANs, they excel in stability and sample quality.\\

\textbf{RGAN.}
Relativistic GAN (RGAN) introduces a relativistic discriminator to enhance the stability and quality of GAN-generated samples \cite{jolicoeur2018relativistic}. Unlike traditional GANs, where the discriminator determines if a sample is real or fake, the RGAN discriminator estimates the probability that a genuine sample is more realistic than a fake sample, and vice versa. It compares the likelihood of a true sample being real with the likelihood of a fake sample being real. This approach guides the generator to produce samples that are more realistic than the discriminator's current estimates for both real and fake samples. To ensure this relativistic nature of RGAN, samples are considered from both real and fake data pairs $\tilde{x}=\left(x_R, x_F\right)$, where $x_R \sim \mathbb{P}_{\text{Real}}$ represents the real data and $x_F \sim \mathbb{P}_{\text{Fake}}$ symbolize its fake counterpart. Mathematically, the generator and discriminator loss functions of the RGAN framework can be expressed as:
\begin{align*}
L_G = &\mathbb{E}_{\left(x_R, x_F\right) \sim\left(\mathbb{P}_{\text{Real}}, \mathbb{P}_{\text{Fake}}\right)}\left[\tilde{g}_1\left(C\left(x_R\right)- C\left(x_F\right)\right)\right] \\ & +
 \mathbb{E}_{\left(x_R, x_F\right) \sim\left(\mathbb{P}_{\text{Real}}, \mathbb{P}_{\text{Fake}}\right)}\left[\tilde{g}_2\left(C\left(x_F\right)-C\left(x_R\right)\right)\right] \text{ and } \\
L_D= & \mathbb{E}_{\left(x_R, x_F\right) \sim\left(\mathbb{P}_{\text{Real}}, \mathbb{P}_{\text{Fake}}\right)}\left[\tilde{f}_1\left(C\left(x_R\right)-  C\left(x_F\right)\right)\right] \\ & +\mathbb{E}_{\left(x_R, x_F\right) \sim\left(\mathbb{P}_{\text{Real}}, \mathbb{P}_{\text{Fake}}\right)}\left[\tilde{f}_2\left(C\left(x_F\right)-C\left(x_R\right)\right)\right], 
\end{align*}
where $C(\cdot)$ is the non-transformed layer and $\tilde{g}_1,\tilde{g}_2, \tilde{f}_1, \tilde{f}_2$ are scalar-to-scalar functions. The term $\left(C\left(x_F\right)-C\left(x_R\right)\right)$ of the modified loss function can be interpreted as the likelihood that the given fake data is more realistic than randomly sampled real data. The relativistic discriminator in RGAN enhances stability by mitigating issues like mode collapse and vanishing gradients, commonly observed in conventional GANs \cite{jolicoeur2018relativistic}. RGAN surpasses regular GANs in generating high-quality samples. It also exhibits improved resilience against adversarial attacks, ensuring sample security. However, these advantages come at the expense of higher computational requirements compared to regular GANs owing to the use of relativistic discriminator \cite{mescheder2017numerics}. Additionally, RGAN necessitates careful hyperparameter tuning, including learning rate and regularization parameters, for optimal performance \cite{gomez2019turbulent, nguyen2016multifaceted, tramer2017ensemble}. Furthermore, the efficacy of RGAN depends on the specific use case, limiting its universal applicability.\\

\textbf{StarGAN.}
StarGAN, a type of GAN model introduced in the work of Choi et al. \cite{choi2018stargan}, is specifically designed for multi-domain image-to-image translations. In contrast to the CycleGAN model \cite{zhu2017unpaired} that focuses on translating images between two specific domains, StarGAN offers the capability to perform translations across a diverse range of domains using a single generator and discriminator. This model trains the generator network $G$ to map the input image $x$ to an output image $y$ conditioned on the randomly generated target domain label $c$ i.e., $G(x, c) \longrightarrow y$. In case of the discriminator network $D$ an additional classifier is used to produce the probability distribution for both source and domain labels $D: x \longrightarrow \{D_{\text{src}(x)}, D_{\text{cls}(x)}\}$. To ensure an efficient multi-domain image translation this framework utilizes several loss functions namely, the adversarial loss, the domain classification loss, and the reconstruction loss. The conventional adversarial loss ensures the generation of high-quality realistic images. The domain classification loss of real images optimizes $D$ to accurately classify $x$ to their input domain label $c^{\prime}$, whereas, the domain classification loss of fake images optimizes $G$ to generate images that can be classified as the generated target domain $c$. Overall, the domain classification loss ensures the coherent multi-domain image classification in the StarGAN model. Furthermore, to ensure that the translated images retain the characteristics of the input image and exclusively modify the domain-related features, a reconstruction loss is used in training the generator network. The overall objective function of the StarGAN model is mathematically expressed as:
\begin{multline*}
    L_G = \mathbb{E}_x\left[\log D_{src}(x)\right]+ \mathbb{E}_{x, c}\left[\log \left(1-D_{s r c}(G(x, c))\right)\right] \\
    - \lambda_1 \mathbb{E}_{x, c}\left[-\log D_{cls}(c \mid G(x, c))\right]\\ + \lambda_2 \mathbb{E}_{x, c, c^{\prime}}\left[\left\|x-G\left(G(x, c), c^{\prime}\right)\right\|_1\right ] \text{ and }\\ 
    L_D = - \mathbb{E}_x\left[\log D_{src}(x)\right] - \mathbb{E}_{x, c}\left[\log \left(1-D_{s r c}(G(x, c))\right)\right] \\
    - \mathbb{E}_{x, c^{\prime}}\left[\log D_{c l s}\left(c^{\prime} \mid x\right)\right],
\end{multline*}
where $\lambda_1$ and $\lambda_2$ are the hyper-parameters that control the effect of the domain classification loss and the reconstruction loss in the StarGAN model, respectively. The training process involves iteratively optimizing the components of the loss functions to achieve high-quality multi-domain image-to-image translations. The StarGAN framework offers several advantages in multi-domain image translation tasks. It utilizes a single generator-discriminator network for all domains, reducing computational complexity. StarGAN can effectively learn domain mappings with limited or unpaired data and preserve the identity of input images in the same target domain. However, it has several drawbacks, including a complex loss function that leads to a time-consuming training process \cite{li2017universal, huang2017arbitrary}. Additionally, regulating image quality and handling translations between complex domains with significant appearance or structural changes can be challenging in StarGAN \cite{isola2017image}. Moreover, this model can be used to manipulate images to a considerable extent which might lead to ethical concerns \cite{thies2016face2face}.\\

\textbf{BigGAN.}
BigGAN, introduced by Brock et al. in 2018, is an innovative methodology for training GAN on a large scale to achieve a high-quality synthesis of natural images \cite{brock2018large}. It aims to address the challenge of generating high-quality images with high resolutions, which traditional GANs struggle to achieve \cite{xia2022gan}. BigGAN stands out by employing large-scale architecture and a unique truncation technique that allows for the generation of high-fidelity images with intricate details and textures. The model is capable of producing images of various resolutions, reaching up to 512 $\times$ 512 pixels, and has been trained on a substantial dataset of images.
Similar to GAN (as in Eq. \ref{GAN_loss}), during the training of BigGAN model gradient descent techniques are used to update the parameters of $G$ and $D$. The discriminator aims to maximize the objective, while the generator aims to minimize it. BigGAN introduces architectural modifications to enhance image quality and diversity. It incorporates class-conditional GANs and self-attention mechanisms. Regularization techniques like orthogonal regularization and truncation tricks stabilize and control the generator's output. Data augmentation methods, such as progressive resizing and interpolation, are employed to handle high-resolution images effectively. The modified training approach in the BigGAN architecture enables the generation of high-quality images with detailed features and textures, surpassing the capabilities of regular GANs. This enhanced model offers scalability, addresses mode collapse issues, and has broad applications in fields such as video processing, satellite imaging, and medical imaging. However, it is computationally demanding, especially when dealing with large datasets or complex models \cite{karras2020training, franceschelli2021creativity}. Additionally, the generalization of the framework to new, unseen data is limited, requiring further fine-tuning or training on fresh datasets \cite{dumoulin2016adversarially}.\\

\textbf{MI-GAN.}
In the field of deep learning, constrained data sizes within the medical domain pose a significant challenge for supervised learning tasks, elevating concerns about overfitting. To address this, Iqbal et al. introduced Medical Imaging GAN (MI-GAN) in 2018, an innovative GAN framework tailored for Medical Imaging \cite{iqbal2018generative}. MI-GAN is specialized in generating synthetic retinal vessel images along with segmented masks based on limited input data. The architecture of the MI-GAN framework's generator network adopts an encoder-decoder structure. Given a random noise vector, the encoder functions as a feature extractor, capturing local and global data representations through its fully connected neural network design. These learned representations are then channeled into the decoder using skip connections, facilitating the generation of segmented images. The generator's enhancements encompass the integration of global standard segmented images and style transfer mechanisms, refining the segmented image generation process. Consequently, the modified MI-GAN generator is trained using a blend of adversarial, segmentation, and style transfer loss functions. In contrast, the discriminator network within the MI-GAN model consists of multiple convolutional layers, and it is trained using adversarial loss functions to effectively distinguish between real and generated images. MI-GAN refines the conditional GAN model for retinal image synthesis and segmentation. Remarkably, despite being trained with a mere ten real examples, this model holds tremendous potential in medical image generation. Nonetheless, this approach relies on spatial alignment to achieve superior outcomes, which can often be scarce \cite{mahmud2021deep}.\\

\textbf{AttGAN.}
AttGAN, also known as Attribute GAN, is a variation of the GAN framework that focuses on generating images with customizable properties such as age, gender, and expression. It was introduced by He et al. in 2019 in their work ``AttGAN: Facial Attribute Editing by Only Changing What You Want'' \cite{he2019attgan}. AttGAN aims to allow users to modify specific facial attributes while preserving the overall identity and appearance of the face. By manipulating attribute vectors, users can control the desired changes in the facial attributes, resulting in realistic and visually appealing image transformations. 
The AttGAN framework combines two subnetworks an encoder $G_{\text{Enc}}$ and a decoder $G_{\text{Dec}}$ in place of $G$ of conventional GAN and it utilizes an attribute classifier $C$ with the discriminator network. During the training phase, given an input image $x^{\tilde{a}}$ with a set of $n$-dimensional binary attribute $\tilde{a}$, $G_{\text{Enc}}$ encodes $x^{\tilde{a}}$ into a latent vector representation i.e., $s = G_{\text{Enc}}\left(x^{\tilde{a}}\right)$. Simultaneously, $G_{\text{Dec}}$ is employed for editing the attributes of $x^{\tilde{a}}$ to another set of $n$-dimensional attributes $\tilde{b}$ i.e., the edited image $x^{\hat{b}}$ is constructed as  $x^{\hat{b}}= G_{\text{Dec}}\left(s, \tilde{b}\right)$. To perform this unsupervised learning task $C$ is used with the encoder-decoder pair to constrain $x^{\hat{b}}$ to possess the desired qualities. Moreover, the adversarial loss used in the training process ensures realistic image generation. On the other hand, to allow for satisfactory preservation of attribute-excluding details in the network a reconstruction loss is utilized in the framework. This loss ensures that the interaction between the latent vector $s$ with attribute $\tilde{b}$ will always produce $x^{\hat{b}}$ and the interaction between $s$ with attribute $\tilde{a}$ will always produce $x^{\hat{a}}$, approximating the input image $x^{\tilde{a}}$. Thus the overall loss function for the encoder-decoder-based generator of AttGAN can be expressed as:
\begin{multline*}
L_{\text{Enc, Dec}} = \lambda_{\text{Rec}} \mathbb{E}_{x^{\tilde{a}}} \left[\|x^{\tilde{a}} - x^{\hat{a}}\|_1\right] \\ + \lambda_{\text{Cls}_G} \mathbb{E}_{x^{\tilde{a}}, \tilde{b}} \left[ \operatorname{H}\left( \tilde{b}, C(x^{\hat{b}}) \right) \right] 
- \mathbb{E}_{x^{\tilde{a}}, \tilde{b}} \left[D\left(x^{\hat{b}}\right)\right]    
\end{multline*}
and the loss for the classifier and the discriminator is formulated as:
\begin{multline*}
    L_{\text{D, Cls}} = \lambda_{\text{Cls}_D} \mathbb{E}_{x^{\tilde{a}}} \left[ \operatorname{H}\left( \tilde{a}, C(x^{\tilde{a}}) \right) \right] - \\ \mathbb{E}_{x^{\tilde{a}}} \left[D\left(x^{\tilde{a}}\right)\right] + \mathbb{E}_{x^{\tilde{a}}, \tilde{b}} \left[D\left(x^{\hat{b}}\right)\right],
\end{multline*}
where $\operatorname{H}$ is the cross entropy loss, and $\lambda_{\text{Rec}}, \lambda_{\text{Cls}_G}, \lambda_{\text{Cls}_D}$ are hyperparameters for balancing the losses. AttGAN offers several benefits in the image generation domain including precise control over the attributes of generated images, allowing users to modify age, gender, expression, and other qualities. It provides flexibility by adapting to multiple domains and tasks, enabling customization and flexibility in image synthesis applications. The model produces realistic images that approximate the desired attributes while maintaining the visual aspects of the original image. However, ethical considerations regarding representation, identity, and privacy must be addressed when using AttGAN or similar models \cite{buolamwini2018gender, dai2022deep}. The computational complexity of AttGAN requires significant resources and may pose challenges for deployment in production settings or on resource-limited devices. Additionally, AttGAN relies on labeled data with attribute annotations, which may not always be readily available, and the performance and generalizability of the model can be influenced by the quantity and quality of the attribute annotations \cite{hou2017deep}. The distribution and diversity of the training data can also impact the model's performance and ability to handle uncommon or out-of-distribution features \cite{reed2016generative}. In conclusion, AttGAN provides precise attribute control, flexibility, and realistic image generation capabilities, but careful ethical considerations, resource requirements, and data dependencies should be taken into account when utilizing the model in practical applications.\\

\textbf{DM-GAN.}
The Dynamic Memory GAN (DM-GAN) introduced by Zhu et al. in 2019 combines the power of GANs with a memory-augmented neural network design to overcome the limitations of conventional GANs \cite{zhu2019dm, li2019diverse}. By addressing issues like mode collapse and lack of fine-grained control, DM-GAN aims to improve the image synthesis process. This deep learning model focuses on generating realistic images from text descriptions, tackling two main challenges in existing methods. Firstly, it addresses the impact of initial image quality on the refinement process, ensuring satisfactory results. Secondly, DM-GAN considers the importance of each word in conveying image content by incorporating a dynamic memory module. The two-stage training of the DM-GAN framework initially transforms the textual description into an internal representation using a text encoder and a deep generator model is utilized to generate an initial image based on the encoded text and random noise. In the subsequent dynamic memory-based image refinement step the generated fuzzy image is processed using a memory writing gate to select relevant text information based on the initial image content and a response gate to fuse information from memories and image features. These advancements enable DM-GAN to generate high-quality images from text descriptions accurately. The dynamic memory module of DM-GAN enhances image generation by capturing long-range relationships and maintaining global context, resulting in persuasive and visually appealing images. It provides fine-grained control over attribute-guided synthesis and increases diversity by addressing mode collapse. However, DM-GAN's computational complexity and memory management pose challenges, and it relies on labeled data \cite{nair2010rectified, bengio1994learning}. The model's interpretability is limited due to the complexity of the memory module \cite{graves2014neural, zeiler2014visualizing}. In conclusion, DM-GAN offers enhanced image generation capabilities with control, diversity, and robustness, while considerations such as computational resources, data availability, and interpretability should be considered.\\

\textbf{SinGAN.}
Single-Image GAN (SinGAN) is an unconditional generative model introduced by Shaham, et al. in 2019 for learning the internal statistics from a single image without the need for additional training data \cite{shaham2019singan}. SinGAN allows for a wide range of image synthesis and manipulation tasks, including animation, editing, harmonization, and super-resolution, among many others. The key innovation of SinGAN is the use of a multi-scale pyramid of GANs, where each GAN is responsible for generating images at a different scale. This hierarchical structure enables SinGAN to capture both the global and local characteristics of the input image, resulting in high-quality and coherent output images. By training on a single image, SinGAN eliminates the need for a large dataset, making it a versatile and practical tool for image generation tasks. During the training phase of SinGAN, a hierarchical structure called the multi-scale pyramid is utilized. This pyramid consists of a series of generators denoted as $\left\{G_0, G_1, \ldots, G_N \right\}$. The generators take input patches of the image at different downsampled levels, represented as $\left\{x_0, x_1, \ldots, x_N\right\}$, where each level is downsampled by a factor of $r^n$ ($r > 1$). The generators, along with their corresponding discriminators $D_n$, are trained using adversarial training. The goal is to generate realistic samples that cannot be distinguished from the downsampled image $x_n$. The SinGAN architecture consists of 5 convolutional blocks in both $G_n$ and $D_n$ networks. Each block consists of a 3$\times$3 convolutional layer with 32 kernels, followed by batch normalization and LeakyReLU activation. The patch size for the discriminator remains fixed at 11$\times$11 across all pyramid levels. During training, the generator and discriminator networks are iteratively updated to optimize a combination of adversarial loss and reconstruction loss. As the training progresses to higher pyramid levels, the generator incorporates the output from the previous level, enabling it to capture finer details and generate more realistic images. To enhance the model's ability to handle diverse variations, noise injection is introduced during training, where random noise patterns are added to the input image at each scale. This helps in generating diverse outputs. The training process continues until convergence, where the generator is capable of synthesizing images that closely resemble the training image at all scales of the pyramid.



SinGAN offers numerous advantages in image manipulation tasks, requiring minimal data. It enables controlled alteration, synthesis, and modification of images, allowing users to adjust lighting, colors, textures, and objects. The model produces aesthetically realistic and visually consistent results that align with the input image. Its multi-stage training process captures global and local characteristics, resulting in high-quality outputs. However, SinGAN lacks explicit control over specific image traits and quality depends on input image quality and quantity \cite{berthelot2018understanding}. Ethical considerations should be addressed, and the model is computationally complex with limited interpretability \cite{brown2020language}. Nevertheless, SinGAN's multi-stage training has gained popularity due to its versatility and the powerful image generation capabilities it offers.\\

\textbf{PATE-GAN.}
In our data-centric world, safeguarding data privacy holds paramount importance, ensuring the protection of individual rights, ethical data handling, and the establishment of a reliable digital environment. It ensures a harmonious blend of leveraging the benefits of data-driven technologies while respecting individual's autonomy and rights. To uphold these concerns and to enable the ethical usage of real-world data in various machine-learning frameworks, Jordan et al. in 2019 proposed the Private Aggregation of Teacher Ensembles Generative Adversarial Network (PATE-GAN) framework \cite{jordon2018pate}. Combining the differential privacy principles of Private Aggregation of Teacher Ensembles (PATE) with the generative prowess of GANs, PATE-GAN generates synthetic data for training algorithms while aiming for a positive societal impact. Similar to the conventional GAN model, PATE-GAN comprises of a generator network that receives a latent vector as input and provides generated data as an output. However, in the discriminator aspect, PATE-GAN innovatively integrates the PATE mechanism involving multiple teacher discriminators and a single student discriminator. The teacher discriminators classify real and generated samples within their dataset segments, while the student discriminator employs the labels aggregated from the teacher discriminators to classify generated samples. The framework's training employs an asymmetric adversarial process, where teachers aim to enhance their loss relative to the generator, the generator targets the student's loss, and the student seeks to optimize its loss against the teachers. This arrangement with the student discriminator ensures differential privacy concerning the original dataset.\\

\textbf{POLY-GAN.}
Introduced by Pandey et al. in 2020, Poly-GAN is a novel conditional GAN architecture aimed at fashion synthesis \cite{pandey2020poly}. This architecture is designed to automatically dress human model images in diverse poses with different clothing items. Poly-GAN employs an encoder-decoder structure with skip connections for tasks like image alignment, stitching, and inpainting. The training procedure of the Poly-GAN framework consists of four steps. This model takes input images, including a reference garment and a model image for clothing placement. Initially, pre-processing involves using a pre-trained LCR-Net$++$ pose estimator \cite{rogez2019lcr} to extract the model's pose skeleton and a U-Net$++$ segmentation network \cite{ronneberger2015u, wang2004image} to obtain the segmented mask of the old garment from the model image. The Poly-GAN pipeline begins by passing the reference garment and generated RGB pose skeleton through the generator to create a garment image that aligns with the skeleton's shape. The architecture of $G$ follows an encoder-decoder structure. The encoder incorporates three components: a Conv module for propagating pose skeleton information at each layer, a ResNet module for generating a feature vector \cite{he2016deep}, and a Conv-norm module with two convolutional layers to process the other two modules' outputs. On the other hand, the decoder learns to produce the desired garment image based on pose condition embedding sent by the encoder using skip connections. The transformed garment image and segmented pose skeleton are sent as inputs to the second stage of the network for image stitching, yielding an image of the pose skeleton with the reference attire. In the third stage, the model performs inpainting to eliminate any irregularities in the generated model image. The discriminator, similar in structure to SR-GAN \cite{ledig2017photo}, is employed during these stages to differentiate real from fake images. Finally, in the fourth stage, post-processing is applied, stitching the model's head to the image to produce the final output. The Poly-GAN framework utilizes adversarial, GAN, and identity losses for training, ensuring high image quality and minimizing texture and color discrepancies from real images. Poly-GAN presents an advancement in fashion synthesis compared to other models \cite{zhu2017your}, as it operates with multiple conditional inputs and achieves satisfactory fitting results without requiring 3D model information \cite{mameli2021deep}. However, the generated images can exhibit texture deformation and body part loss, affecting the fitting outcomes \cite{wu2022design}. Further research is needed to address these issues in this domain.\\

\textbf{MIEGAN.}
Mobile Image Enhancement GAN (MIEGAN), introduced by Pan et al. in 2021, is a novel approach within the realm of GAN-based architectures, with the primary objective of elevating the visual caliber of images taken via mobile devices \cite{pan2021miegan}. This endeavor involves several modifications to the conventional GAN architecture. In the MIEGAN model, a multi-module cascade generative network is utilized which combines an Autoencoder and a feature transformer. The encoder of this modified generator comprises of two streams with the second stream being responsible for enhancing the regions with low luminance - a common issue in mobile photography leading to reduced clarity. In the feature transformative module, the local and global information of the image is further captured using a dual network structure. Furthermore, to enhance the generative network's ability to produce images of superior visual quality, an adaptive multi-scale discriminator is employed in lieu of a standard single discriminator in the MIEGAN model. This multi-scale discriminator serves to differentiate between real and fake images on both global and local scales. To harmonize the evaluations from the global and local discriminators, an adaptable weight allocation strategy is utilized in the discriminator. Additionally, this model is trained based on a contrast loss mechanism and a mixed loss function, which further enhances the visual quality of the generated images. Despite the image quality enhancement capabilities of the MIEGAN framework, their high computation complexity poses a significant challenge for their real-time application in mobile photography.\\

\textbf{VQGAN.}
Vector Quantized GAN (VQGAN) introduces a novel methodology that merges the capabilities of GAN with vector quantization techniques to generate high-quality images \cite{esser2021taming}. This approach effectively leverages the synergies between the localized interactions of CNN and the extended interactions of Transformers \cite{vaswani2017attention} in tasks involving the conditional synthesis of data. The distinctive architecture of VQGAN not only yields images of exceptional quality but also empowers a degree of creative influence, enabling the manipulation of various attributes within the generated content. The training process of the VQGAN architecture unfolds in two pivotal phases. Initially, a variational autoencoder and decoder are trained, as opposed to the conventional GAN generator network. This training aims to reconstruct the image by utilizing a discrete latent vector representation derived from the input image. This intermediate representation is subsequently linked to a codebook, efficiently capturing the underlying semantic information. To augment the fidelity of the reconstructed image, a discriminator is incorporated into the autoencoder structure. The training of the autoencoder model, the codebook, and the discriminator involves optimizing a fusion of adversarial loss and perceptual loss functions. In the subsequent phase, the codebook indices, constituting the intermediate image representations, are fed into Transformers. These Transformers are trained through a transformer loss mechanism, guiding them to predict the succeeding indices within the encoded sequence, resulting in an improved codebook representation. Finally, the information from the codebook is utilized by the decoder to generate images of higher resolutions. The unique aspect of VQGAN lies in its ability to allow users to manipulate generated images in creative ways. By modifying the quantized codes, users can control specific features of the generated content, thereby unlocking a spectrum of artistic potentials. Nonetheless, the caliber of the images generated by VQGAN depends largely on its input data, necessitating expansive datasets and substantial computational resources to produce images of exceptional excellence \cite{chaitanya2023local}. Consequently, this restricts its immediate applicability in real-time case studies. Moreover, the codebook representation used in the vector quantization process can significantly reduce the variation in the generated images \cite{kalchbrenner2017video}.\\

\textbf{DALL-E.}
DALL-E is an advanced text-to-image generative framework created by OpenAI that utilizes a two-stage process to generate images from textual prompts \cite{radford2021dalle, ramesh2021zero}. It combines the concepts of GANs and Transformers to generate highly realistic and coherent images from textual descriptions. What sets DALL-E apart is its ability to generate realistic art and images from textual descriptions that may describe completely novel concepts or objects. The working principle of the pre-trained DALL-E model comprises of two phases. The first stage involves a prior model that generates a Contrastive Language-Image Pretraining (CLIP) \cite{radford2021learning} image embedding, capturing the essential gist of the image based on the provided caption. In the second stage, a decoder model known as GLIDE takes the image embedding and reconstructs the image itself, gradually removing noise and generating a realistic and visually coherent image. The CLIP model, consisting of a text encoder and an image encoder, is trained using contrastive training to learn the relationship between images and their corresponding captions. This allows the model to generate the CLIP text embedding from the input caption. Further, the prior model of DALL-E processes this text representation to generate the CLIP image embedding. In case of the decoder, DALL-E utilizes a Diffusion model \cite{sohl2015deep} which generates the image by using CLIP image embedding and the CLIP text embedding as an additional input. DALL-E's two-stage process offers advantages in prioritizing high-level semantics and enabling intuitive transformations. It excels in generating creative and imaginative images based on textual descriptions, making it valuable for creative tasks. However, training DALL-E requires substantial computational resources and presents challenges in fine-tuning and attribute control. Ethical concerns and biases surrounding AI-generated content also arise \cite{singh2021illiterate, marcus2022very}. Moreover, the lack of interpretability and explainability of this framework restricts its applications in legal, medical, or safety-sensitive domains \cite{rudin2019stop}. Nevertheless, DALL-E represents a significant advancement in image synthesis and has garnered attention for its creative potential. Ongoing research, such as DALL-E 2 \cite{ramesh2022hierarchical}, continues to push the boundaries of this field and attempts to mitigate the explainability concerns \cite{doshi2017towards}.\\

\textbf{CEGAN.}
Class imbalance is a prevalent challenge across many real-world datasets. In the context of classification tasks, this skewed distribution of classes leads to a significant bias favoring the majority class. Previous studies have suggested oversampling approaches, involving the artificial generation of samples from the minority class, as an efficient mechanism to mitigate this issue. Classification Enhancement GAN (CEGAN) model introduces a solution to address the class imbalance issue through the utilization of a GAN-based framework, as outlined in the work by Suh et al. \cite{suh2021cegan}. This model particularly focuses on enhancing the quality of data generated from the minority class, thereby mitigating the classifier's bias towards the distribution of the majority class. Differing from the conventional GAN model, the CEGAN framework combines three distinct networks -- a generator, a discriminator, and a classifier. The training process of the CEGAN model involves a two-step sequence. In the initial phase, the generator generates synthetic data using input noise and real class labels. Simultaneously, the discriminator distinguishes between real and synthetic data, while the classifier assigns class labels to input samples. The subsequent stage involves the integration of the generated samples with the original training data, creating an augmented dataset for training the classifier. The CEGAN framework serves as an efficient methodology that incorporates techniques such as data augmentation, noise reduction, and ambiguity reduction to effectively tackle class imbalance problems. Notably, this approach overcomes the limitations associated with traditional resampling techniques, as it avoids the need to modify the original dataset. \\

\textbf{SeismoGen.}
Seismogen is a seismic waveform synthesis technique that utilizes GAN for seismic data augmentation \cite{wang2021seismogen}. The motivation behind Seismogen arises from the need for abundant labeled data for accurate earthquake detection models. To overcome the scarcity of seismic waveform datasets, Wang et al. introduced the Seismogen framework, employing GAN to generate realistic multi-labeled waveform data based on limited real seismic datasets. Incorporating this additional dataset enhances the training of machine learning-based seismic analysis models, leading to more robust predictions for out-of-sample datasets. The mathematical formulation of the Seismogen framework follows the Wasserstein GAN \cite{arjovsky2017wasserstein} framework and can be expressed as:
\begin{align*}
L_G=&-\underset{z \sim \operatorname{N}(0,1)}{\mathbb{E}} D(G(z)),\\
L_D=&\underset{z \sim \operatorname{N}(0,1)}{\mathbb{E}} D(G(z))-\underset{x \sim p_{\text{data}}}{\mathbb{E}} D(x) \\ &+\lambda \underset{z \sim \operatorname{N}(0,1)}{\mathbb{E}}\left[\left(\|D(G(z))\|_2-1\right)^2\right],
\end{align*}
where the noise $z$ is a standard normal variable and $\lambda$ is a hyperparameter. The primary objective is to minimize the difference between the true seismic waveforms and the synthetic waveforms generated by the Seismogen. This is achieved by iteratively optimizing $L_G$ and $L_D$ to find an equilibrium between the generator and discriminator networks. SeismoGen has demonstrated its ability to generate highly realistic seismic waveforms, making it valuable for seismic waveform analysis and data augmentation. Its conditional generation feature allows users to produce waveforms labeled with specific categories, enhancing its versatility for various applications. SeismoGen is scalable and capable of generating large databases of artificial waveforms, which is beneficial for tasks requiring extensive training data. However, SeismoGen's effectiveness is influenced by the quality and distribution of the training data. It does not model the expected waveform move-out, which is relevant in various seismic research. Additionally, due to imbalanced real seismic waveform datasets, SeismoGen struggles to generate data with rare characteristics. Moreover, the computational cost of training and using SeismoGen may be a limiting factor, especially for real-time seismic hazard assessment applications. As a relatively new technology, there might be some potential for unexpected behavior when using SeismoGen, as its full capabilities and limitations are yet to be fully explored.\\

\textbf{MetroGAN.}
Zhang et al. introduced Metropolitan GAN (MetroGAN) as a geographically informed generative deep learning model for urban morphology simulation \cite{zhang2022metrogan}. MetroGAN incorporates a progressive growing structure to learn urban features at various scales and leverages physical geography constraints through geographical loss to ensure that urban areas are not generated on water bodies. The generation of cities with MetroGAN involves a global city dataset comprising three layers: terrain (digital elevation model), water, and nighttime lights, effectively capturing the physical geography characteristics and socioeconomic development of cities. The model detects and represents over 10,000 cities worldwide as 100km $\times$ 100km images. The mathematical formulation of the MetroGAN framework is a modified version of the LSGAN model \cite{mao2017least}, which can be expressed as follows: 
\begin{multline*}
L^* = \operatorname{arg}\underset{G}{\operatorname{min}}\;\underset{D}{\operatorname{max}} \frac{1}{2} \mathbb{E}_{x,y}\left[\left(D(x,y) - 1 \right)^2 \right] \\ 
+  \frac{1}{2} \mathbb{E}_{x,z}\left[\left(D(x,G(x,z))\right)^2 \right] 
+ \lambda_{L1} L_{L1}(G)\\ - \lambda_{\text{Geo}} \mathbb{E}_{x,z}\left[x_{\text{water}} \odot G(x, z)\right],
\end{multline*}
where images $x$ with corresponding labels $y$ and a random vector $z$ in the latent space are fed into $G$ to produce simulated images $G(x, z)$. Both real input pairs $(x, y)$ and simulated pairs $(x, G(x, z))$ are then presented to $D$ to distinguish real images from fake ones and also to assess if the input pairs match. The objective loss function comprises different terms, including least square adversarial loss (from the first two expectation terms), $L1$ loss denoted as $L_{L1}$, and a geographical loss with hyperparameters $\lambda_{L1}$ and $\lambda_{\text{Geo}}$, respectively. The geographical loss (last term) utilizes Hadamard product $\odot$ to filter out pixels that generate urban areas on water area $x_{\text{water}}$. MetroGAN, a robust urban morphology simulation model, has several notable advantages and limitations. On the positive side, it incorporates geographical knowledge, resulting in enhanced performance. Its progressive growing structure allows for stable learning at different scales, while multi-layer input ensures precise city layout generation. The model's evaluation framework covers various aspects, ensuring the quality of its output. Furthermore, MetroGAN finds wide applications in urban science and data augmentation. However, these strengths come with challenges, including high computational costs due to extensive data requirements and dependence on data quality, which may hinder its performance with noisy or missing data. Additionally, the model lacks interpretability, making it difficult to understand the reasoning behind its predictions, and it may struggle to represent all intricate features of complex urban systems effectively.\\

\textbf{M3GAN.}
Anomaly detection in multi-dimensional time series data has received tremendous attention in the fields of medicine, fault diagnosis, network intrusion, and climate change. In this work, the authors have proposed the M2GAN (a GAN framework based on a masking strategy for multi-dimensional anomaly detection) and M3GAN (M2GAN for mutable filter) for improving the robustness and accuracy of GAN-based anomaly detection methods. M2GAN generates fake samples by directly reconstructing real samples, which are sufficiently realistic \cite{li2023m3gan}. This is done by extracting various information from the original data by the mask method which improves the robustness of the model. M3GAN fuses the fast Fourier transform (FFT) \cite{brigham1988fast} and wavelet decomposition \cite{percival2000wavelet} to obtain a mutable filter to process the raw data so that the model can learn various types of anomalies. The architecture of the M2GAN framework utilizes the AAE \cite{makhzani2015adversarial} in place of the generator of the conventional GAN model for generating realistic fake data. A masking strategy of the AAE enhances the variability within the original time series and overcomes the mode collapse problem. For the discriminator network, this framework employs an AnoGAN \cite{schlegl2017unsupervised} architecture that distinguishes between normal data and anomalous data using DCGAN \cite{radford2015unsupervised}. The M3GAN model combines a dynamic switch-based adaptive filter selection mechanism with the multidimensional anomaly detection capabilities of the M2GAN model. This approach allows one to select the most suitable filter for the given data that better exploits the complex characteristics of the series, leading to improved accuracy in anomaly detection. Both M2GAN and M3GAN architectures excel in spotting anomalies in multi-dimensional time series data, offering adaptability for dynamic settings. Its capacity to generate synthetic data aids tasks like diverse model training. However, their high computational complexity leads to extended processing times. Moreover, their limited interpretability also poses a significant challenge in understanding the marked anomalies. Further research is needed in this domain to address these issues and provide support for adaptive filter parameters in M3GAN.\\

\textbf{CNTS.}
Cooperative Network for Time Series (CNTS), introduced by Yang et al. in 2023, is a reconstruction-based unsupervised anomaly detection technique for time series data \cite{yang2023cnts}. This model aims to overcome the limitations of the previous generative methods that were sensitive to outliers and showed sub-optimal anomaly detection performance due to their emphasis on time series reconstruction. The CNTS framework consists of two FEDformer \cite{zhou2022fedformer} networks, namely a reconstructor ($R$) and a detector ($D$). The reconstructor aims to regenerate the series that closely matches the known data distribution (without anomalies) i.e., data reconstruction. On the other hand, the detector focuses on identifying the values that deviate from the fitted data distribution, effectively detecting anomalies. Despite having different purposes, these two networks are trained using a cooperative mode, enabling them to leverage mutual information. During the training phase, the reconstruction error of $R$ serves as a labeling mechanism for $D$, while $D$ provides crucial information to $R$ regarding the presence of anomalies, enhancing the robustness to outliers. Thus the multi-objective function of the CNTS model can be expressed as:
$$\left[\begin{array}{c}
         \underset{\theta_D \theta_R}{\operatorname{min}} \sum_{i=1}^n L_D (D(x_i, \theta_D), L_R(x_i, R(x_i, \theta_R))) \\
    \underset{\theta_D \theta_R}{\operatorname{min}} \sum_{i=1}^n (1 - \hat{y}_i(x_i, \theta_D)) L_R(x_i, R(x_i, \theta_R)))
    \end{array}\right],$$
where $x_i$ is the value for the $i^{th}, i = 1, 2, \ldots, n$ time stamp of the input series, $\theta_D$ and $\theta_R$ denotes the parameters of $D$ and $R$, while $L_D$ and $L_R$ represent their corresponding loss functions, respectively. The categorical label $\hat{y}_i$ indicates the presence of anomalies as identified by $D$ and helps to remove data with high anomaly scores, thereby reducing their impact on the training of $R$.
The cooperative training approach employed by CNTS allows it to model complex temporal patterns present in real-world time series data, thus significantly enhancing its performance in various anomaly detection tasks. The flexibility and adaptability of the CNTS model make it robust to the presence of outliers in the series. However, the presence of the dual-network architecture of the CNTS model increases its computational complexity, hindering its real-time applicability. Moreover, the lack of interpretability of the model poses a significant challenge to its potential use cases. Furthermore, the success of the CNTS model is contingent on the availability of representative and diverse time series datasets and the choice of sub-networks. Further research in this domain is required to comment on the performance of the model for diverse datasets and appropriate sub-network choices.\\

\textbf{RidgeGAN.}
RidgeGAN, introduced by Thottolil et al. in 2023, is a hybridization of the nonlinear kernel ridge regression (KRR) \cite{vovk2013kernel, murphy2012machine} and the generative CityGAN model \cite{thottolil2023prediction}. This framework aims to predict the transportation network of the future small and medium-sized cities of India by analyzing the spatial indicators of human settlement patterns. This prediction is crucial for facilitating sustainable urban planning and traffic management systems. The RidgeGAN framework operates in three steps. Firstly, it generates an urban universe for India based on spatial patterns by learning urban morphology using the CityGAN model \cite{Albert2018ModelingUP}. Secondly, it utilizes KRR to study the relationship between the human settlement indices (HSI) and the transportation indices (TI) of 503 real small and medium-sized cities in India. Finally, the KKR model's regression framework is applied to the synthetic hyper-realistic samples of future cities and their TI is predicted. RidgeGAN framework has its applications in diverse areas, such as analyzing urban land patterns, forecasting essential urban infrastructure, and assisting policymakers in achieving a more inclusive and effective planning process. Moreover, this model is especially valuable when designing the transportation network of developing nations with limited or partial real data, as the model can produce data that closely resembles actual urban morphology and helps in data augmentation. However, the framework fails to showcase its performance for the generated human settlements which is crucial in the urban planning procedure. Further studies in this domain are indeed required to understand the suitability of the framework for large cities as well.

\begin{table*}[!t]
\caption{Software Links for the GANs}
\label{tab:software-libraries}
\centering
\begin{tabular}{|l|l|l|l|l|r|}
\hline
\textbf{Index} & \textbf{Software name} & \textbf{Language} & \textbf{Backend}  & \textbf{Link} &\textbf{Ref.}\\
\hline
1 & {CGAN}  & Python & PyTorch & \url{https://github.com/Lornatang/CGAN-PyTorch}& \cite{mirza2014conditional}  \\ \hline
2 & {DCGAN}  & Python  & PyTorch & \url{https://github.com/Natsu6767/DCGAN-PyTorch}& \cite{radford2015unsupervised,goodfellow2016nips,goodfellow2014generative}\\ \hline
3 & {AAEs}  & Python & TensorFlow &  \url{https://github.com/conan7882/adversarial-autoencoders}& \cite{makhzani2015adversarial}  \\ \hline
4 & {InfoGAN}  & Python & TensorFlow & \url{https://github.com/openai/InfoGAN}& \cite{chen2016infogan} \\ \hline
5 & {SAD-GAN}  & -- & -- & -- & \cite{ghosh2016sad} \\ \hline
6 & {LSGAN}  & Python & PyTorch & \url{https://github.com/xudonmao/LSGAN}& \cite{mao2017least} \\ \hline
7 & {SRGAN}  & Python & TensorFlow &  \url{https://github.com/tensorlayer/SRGAN}& \cite{ledig2017photo,tensorlayer2017,tensorlayer2021}\\ \hline
8 & {WGAN}  & Python & PyTorch & \url{https://github.com/Zeleni9/pytorch-wgan}& \cite{arjovsky2017wasserstein,gulrajani2017improved} \\ \hline
9 & {CycleGAN}  & Python & TensorFlow & \url{https://github.com/junyanz/CycleGAN}& \cite{CycleGAN2017,zhu2017unpaired} \\ \hline
10 & {ProGAN}  & Python & PyTorch & \url{https://github.com/tkarras/progressive_growing_of_gans}& \cite{karras2017progressive} \\ \hline
11 & {MidiNet} & Python & TensorFlow & \url{https://github.com/RichardYang40148/MidiNet} & \cite{yang2017midinet} \\
\hline
12 & {SN-GAN} & Python & PyTorch & \url{https://github.com/hanyoseob/pytorch-SNGAN}& \cite{miyato2018spectral} \\ \hline
13 & {RGAN} & Python & TensorFlow & \url{https://github.com/ratschlab/RGAN}& \cite{esteban2017real,jolicoeur2018relativistic} \\ \hline
14 & {StarGAN} & Python & PyTorch & \url{https://github.com/yunjey/stargan}& \cite{choi2018stargan} \\ \hline
15 & {BigGAN} & Python & PyTorch & \url{https://github.com/ajbrock/BigGAN-PyTorch}& \cite{brock2018large} \\ \hline
16 & {MI-GAN} & Python & TensorFlow & \url{https://github.com/hazratali/MI-GAN} & \cite{iqbal2018generative} \\
\hline
17 & {AttGAN} & Python & TensorFlow & \url{https://github.com/LynnHo/AttGAN-Tensorflow} & \cite{he2019attgan,zhang2018generative}\\ \hline
18 & {PATE-GAN} & Python & TensorFlow & \url{https://github.com/vanderschaarlab/mlforhealthlabpub/tree/main/alg/pategan} & \cite{jordon2018pate} \\ \hline
19 & {DM-GAN} & Python & PyTorch & \url{https://github.com/MinfengZhu/DM-GAN}& \cite{zhu2019dm}\\ \hline
20 & {SinGAN} & Python & PyTorch & \url{https://github.com/tamarott/SinGAN} & \cite{shaham2019singan}\\ \hline
21 & {POLY-GAN} & Python & PyTorch & \url{https://github.com/nile649/POLY-GAN} & \cite{pandey2020poly} \\ \hline
22 & {MIEGAN} & -- & -- & -- & \cite{pan2021miegan} \\ \hline
23 & {VQGAN} & Python & PyTorch &  \url{https://github.com/dome272/VQGAN-pytorch}& \cite{esser2021taming,razavi2019generating}\\ \hline
24 & {DALL-E} & Python & PyTorch &  \url{https://github.com/lucidrains/DALLE-pytorch}& \cite{ramesh2021zero,radford2021dalle}\\ \hline
25 & {CEGAN} & -- & -- & -- & \cite{suh2021cegan}\\ \hline
26 & {Seismogen} & Python & PyTorch & \url{https://github.com/Miffka/seismogen} & \cite{wang2021seismogen} \\ \hline
27 & {MetroGAN} & Python & PyTorch & \url{https://github.com/zwy-Giser/MetroGAN}& \cite{zhang2022metrogan} \\
\hline
28 & {M3GAN} & Python & PyTorch & \url{https://github.com/SLZWVICTOR/M3GAN} & \cite{li2023m3gan} \\
\hline
29 & {CNTS} & Python & PyTorch & \url{https://github.com/BomBooooo/CNTS/tree/main} & \cite{yang2023cnts} \\
\hline
30 & {RidgeGAN} & Python & PyTorch & \url{https://github.com/rahisha-thottolil/ridgegan} & \cite{thottolil2023prediction} \\
\hline
\end{tabular}
\end{table*}

\section{Recent Theoretical Advancements of GAN}\label{Section_Theory}
Empirical studies have shown great success of GAN and their variants in producing state-of-the-art results in diverse domains ranging from image, video, and text generation to automatic vehicles, time series, and drug discovery, among many others. The mathematical reasoning of GANs is to approximate the unknown distribution of a given data by optimizing an objective function through an adversarial game between a family of generators and a family of discriminators. Biau et al. \cite{biau2020some} analyzed the mathematical and statistical properties of GANs by establishing connections between adversarial principles and Jensen-Shannon (JS) divergence. Their work provides the large sample properties for the parameters of the estimated distribution and a result towards the central limit theorem. Another cousin approach of GAN called WGAN has more stable training dynamics than typical GANs. Biau et al. \cite{biau2021some} studied the convergence of empirical WGANs when sample size approaches infinity. More recently, the rate of convergence for density estimation with GANs has been studied in \cite{belomestny2021rates}. In particular, they studied the non-asymptotic properties of the vanilla GAN and derived a theoretical guarantee of the density estimation with GANs under a proper choice of deep neural network classes representing generators and discriminators. It suggests that the resulting estimates converge to the true density ($p^*$) in terms of the JS divergence at the rate of $\left(\log n/n\right)^{2\beta/\left(2\beta+d\right)}$, where $n$ is the sample size, $\beta$ determines the smoothness of $p^*$, and $d$ is the data dimension. In Theorem 2 of \cite{belomestny2021rates} if the choice of $G$ and $D$ to be classes of neural networks with rectified quadratic unit (ReQU) activation functions, the rates of convergence for the estimate $p_{\hat{g}}$ to the true density $p^*$ in terms of JS divergence holds the following inequality with probability at least $1-\delta$;
$$
    \operatorname{JS}\left(p_{\hat{g}}, p^*\right) \lesssim \left(\frac{\log n}{n}\right)^{\frac{2\beta}{2\beta+d}} + \frac{\log \left(1/\delta\right)}{n}.
$$
The above mathematical result suggests that the convergence rate of vanilla GAN's density estimate in the JS divergence is faster than $n^{-1/2}$ when $\beta > \frac{d}{2}$; therefore, the obtained rate is minimax optimal for the considered class of densities. Meitz et al. \cite{meitz2021statistical} studied statistical inference for GAN by addressing two critical issues for the generator and discriminator's parameters, namely consistent estimation and confidence sets. Mbacke et al. \cite{mbacke2023pac} studied PAC-Bayesian generalization bound for WGANs based on Wasserstein distance and Total variational distance. The generalization properties of GANs try to answer the following question: How to certify that the learned distribution $p_{\hat{g}}$ is ``close'' to the true one $p^*$? This question is pivotal since the true distribution $p^*$ is unknown in real problems and generative models can only access its empirical counterpart. Liu et al. \cite{liu2017approximation} studied how well GAN can approximate the target distribution under various notions of distributional convergence. Lin et al. \cite{lin2021privacy} showed that under certain conditions GAN-generated samples inherently satisfy some (weak) privacy guarantees. Another study offers a theoretical perspective on why GANs sometimes fail for certain generation tasks, in particular, sequential tasks such as natural language generation \cite{alvarez2022gans}. Further research on the comparative theoretical aspects, both pros and cons, of different generative approaches will enhance support for the wide applications of GANs and address their limitations.

\section{Evaluation Measures}\label{Evaluation}

In contrast to conventional deep learning architectures that employ convergence-based optimization of the objective function, generative models like GANs utilize a minimax loss function, trained iteratively to establish equilibrium between the generator and discriminator networks \cite{goodfellow2014generative}. The absence of an objective loss function for GAN training restricts the ability of loss measurements to assess training progress or model performance. To address this challenge, a mix of qualitative and quantitative GAN evaluation approaches has been developed \cite{borji2019pros}. These evaluation measures particularly vary based on the quality and diversity of the generated synthetic data, as well as the potential applications of the generated data \cite{xu2018diversity}.

Owing to the lack of consensus amongst the researchers on the use of a universal metric to gauge the performance of the deep generative models, different metrics have been developed in the last decade with their unique strengths and particular applicability \cite{goodfellow2016nips}. In this section, we will briefly overview the popular evaluation measures used in different applications.

\subsection{Inception Score}
The Inception Score (IS) is a widely used metric to assess the quality and diversity of GAN-generated samples \cite{salimans2016improved}. It leverages a pre-trained neural network classifier called Inception v3 \cite{szegedy2016rethinking}, which was initially trained on the Imagenet \cite{deng2009imagenet} dataset containing a diverse range of real-world images categorized into 1,000 classes. The IS measures the quality of generated samples based on their classification probabilities predicted by Inception v3. Essentially, higher-quality samples are expected to be strongly classified into specific classes, implying low entropy. In general, the IS value ranges between 1 and the number of classes in the classifier, reflecting the diversity of the generated samples, with higher scores indicating better performance. Nevertheless, the Inception Score does come with a number of limitations. It encounters challenges when dealing with instances of mode collapse, wherein the generated samples by GANs are extremely similar, causing artificially inflated IS values that don't accurately represent diversity. Additionally, it relies on the performance of the Inception v3 model, which might not always align with human perception of image quality. To mitigate these drawbacks of IS, several modified versions have been proposed in the literature. For example, the modified Inception Score (m-IS) attempts to address the mode collapse problem in GAN by evaluating the diversity of images with the same category \cite{gurumurthy2017deligan}. Other modification of IS includes the Mode Score (MS) which evaluates the quality and diversity of the generated data by considering the prior data distribution of the labels \cite{nowozin2016f}.

\subsection{Fréchet Inception Distance}

The Fréchet Inception Distance (FID) is a widely used evaluation metric that measures the quality and diversity of GAN-generated images \cite{heusel2017gans}. It calculates the similarities and differences between the distributions of real and generated images using the Fréchet distance, which is a form of the Wasserstein-2 distance. The FID metric calculates the mean and covariance of both the real and generated images and then computes the distance between their distributions. Mathematically the FID is expressed as:
$$
\operatorname{FID} =\left|\mu-\mu_w\right|^2+\operatorname{tr}\left(\Sigma+\Sigma_w-2\left(\Sigma \Sigma_w\right)^{1 / 2}\right),
$$
where ($\mu$, $\Sigma$) and ($\mu_w$, $\Sigma_w$) represent the mean and covariance pair for the real images and the generated images respectively.

The strength of FID lies in its ability to account for various forms of contamination, such as Gaussian noise, Gaussian blur, black rectangles, and swirls, among others. FID's incorporation of these factors contributes to a more robust evaluation of GAN-generated images. As a widely accepted and utilized metric, FID offers a common ground for comparing results across different GAN architectures, promoting a standardized approach for assessing image quality \cite{karras2017progressive, karras2019style, daras2020your}. 

\subsection{Multi-Scale Structural Similarity}

The Multi-Scale Structural Similarity metric (MS-SSIM), an extension of the traditional Structural Similarity Index (SSIM), serves as an effective measure for evaluating the quality of GAN-generated images \cite{wang2003multiscale}. MS-SSIM focuses on comparing image structures, including luminance and contrast, across different scales. This metric provides a comprehensive evaluation of the similarity between the real and synthesized datasets, considering their structural and geometric aspects. Moreover, the ability of MS-SSIM to account for strong dependencies between closely correlated pixels enhances its sensitivity to perceptual quality.

\subsection{Classifier Two-Sample Test}

Classifier Two-Sample Test (C2ST) is a classification-based approach that evaluates the generalization capabilities of GAN for any synthetic data generation task \cite{lehmann1986testing}. This metric utilizes a classifier (for example, 1-Nearest Neighbour \cite{cunningham2021k}) to distinguish between the real and generated samples. The performance of this classifier is then used as a metric to determine the quality of the generated samples. 
The C2ST metric provides an essential tool for measuring the performance of GAN-based architectures for any applied domains, since the classifier is not restricted to a specific data type. Moreover, it focuses on the discriminative aspect of the generated data quality and complements other evaluation metrics that focus on the distributional and perceptual aspects of the generated data.

\subsection{Music Evaluation Metric}

Evaluating the quality of music generated by GANs presents unique challenges due to the subjective nature of musical perception. Traditional quantitative metrics like those used for image evaluation may not fully capture the richness and complexity of musical content. However, several methods have been developed to assess the quality and coherence of GAN-generated music. Certain objective evaluation metrics encompass factors such as musical characteristics, structure, style, uniqueness, and tonality, drawing from statistical representations \cite{ji2023survey}. Amid these, subjective listening is the most reliable metric for evaluating GAN-generated music. This approach encompasses dimensions like melody, harmony, rhythm, and emotional resonance, thereby furnishing insightful glimpses into the musical caliber.

\subsection{Maximum Mean Discrepancy}
Maximum Mean Discrepancy (MMD) is a statistical measure that quantifies the dissimilarity between two probability distributions. In the context of GAN evaluation, MMD is employed to assess the quality of generated samples by comparing them with real data distributions based on their mean values in a high-dimensional space \cite{bounliphone2015test}. A lower MMD score indicates that the difference between the two data distributions is relatively smaller, hence the synthetic data is similar to the original data. 

\subsection{Time Series Evaluation Metric}
Assessing time series GAN models presents a notable challenge due to the temporal dependencies inherent in the data. Traditional evaluation metrics tailored to static image datasets struggle to capture the intricate patterns found in sequential data. As a result, a combined approach of qualitative and quantitative measures is employed for evaluation purposes \cite{brophy2023generative}. Qualitative assessment relies primarily on human visual judgment when examining the generated samples. However, these methods lack objectivity. To address this limitation, a range of quantitative evaluation techniques is employed within GAN-based time series evaluation. These encompass metrics such as root mean square error, Wasserstein-1 distance, dynamic time warping, and Pearson correlation coefficient, among others.

\subsection{Uncertainty Quantification in GANs}
Uncertainty Quantification (UQ) plays a vital role in characterizing and estimating the uncertainties in both computation and real-world applications. Due to the fact that the analysis of physical processes based on computer models is riddled with uncertainty, therefore, it has to be addressed to perform `trustworthy' model-based inference \cite{volodina2021importance}. Oberdiek et al. presented a method to quantify uncertainties of deep neural networks in image classification based on GANs. By employing GANs to generate out-of-distribution (OoD) samples, their methodology enables the classifier to effectively gauge uncertainties for both OoD examples and minor positives \cite{oberdiek2022uqgan}. He et al. presented a survey on UQ models for deep neural networks based on two types of uncertainty sources, namely data uncertainty and model uncertainty \cite{he2023survey}. They highlighted that GAN-based models can capture the structure of data uncertainty, however, they are hard to train. Another survey \cite{gawlikowski2023survey} highlighted various measures to quantify uncertainties in deep neural networks. However, it still remains difficult to validate existing methods due to the lack of uncertain ground truths.

\section{Limitations and scope for improvement}\label{Section_Limitations}
Although GANs have brought a transformative shift in generative modeling, it's crucial to address the substantial challenges embedded within their training process that demand careful consideration \cite{salimans2016improved}. Various architectural modifications of GAN (as discussed in Section \ref{gan_vARIANTS_SECTION}) aim to address specific GAN-related issues and optimize their overall performance. In this section, we summarize the different obstacles in GAN and discuss their potential remedies.

\subsection{Mode Collapse}
The foremost challenge during GANs training is mode collapse (MC), a phenomenon where the generator's output becomes constrained, yielding repetitive samples that lack the comprehensive range of the target data distribution \cite{ramesh2021zero}. MC arises when the generator doesn't explore the full spectrum of potential outputs and instead generates identical outputs for distinct inputs from the latent space. This issue can manifest due to an overpowering discriminator or insufficient feedback for the generator to diversify its outputs \cite{samangouei2018defense}. Partial and complete mode collapse are its two variants, with the former leading to a limited diversity in generated data and the latter resulting in entirely uniform patterns across generated samples. While partial mode collapse is common, complete mode collapse is relatively rare \cite{goodfellow2016nips}.

Many efforts have been made to tackle the mode collapse problem \cite{de2021bures, li2021tackling}. Some of these approaches include the application of Unrolled GAN \cite{metz2016unrolled} where the generator network is updated by unrolling the discriminator's update steps, unlike the conventional GAN, where $D$ is first updated while $G$ is kept fixed and $G$ is updated based on the updated $D$. Moreover, mini-batch discrimination is often used to mitigate the MC problem \cite{salimans2016improved}. In this approach, instead of modeling each data example independently, $D$ processes multiple data examples in mini-batches. The use of modified loss functions, for example, Least-Square GAN \cite{mao2017least}, Wasserstein GAN \cite{arjovsky2017wasserstein}, Cycle consistency GAN \cite{zhu2017unpaired} also reduces the mode collapse problem.

\subsection{Vanishing Gradients}

The vanishing gradients problem is another significant challenge encountered during the training phase of GANs. This issue emerges due to the complex architecture of GANs, where both $G$ and $D$ need to maintain a balance and learn collaboratively \cite{zhang2019towards}. During the training process, as gradients are backpropagated through the layers of the network, they can diminish drastically, leading to stagnancy in learning. This circumstance can occur when the discriminator becomes very accurate, such as when $D(G(z) = 0 \text{ and } D(x) = 1$ or when $D$ is inadequately trained and fails to differentiate between real and generated data. Consequently, the loss function might approach zero, hindering constructive feedback to the generator and restricting the generation of high-quality data. Several strategies have been proposed to address vanishing gradients in GANs. One approach is to use a modified loss function, such as the Least-Square GAN \cite{mao2017least} that mitigates the vanishing gradient problem to a considerable extent. Furthermore, advanced optimization algorithms, alternative activation functions, and batch normalization strategies are often adopted to reduce the effect of vanishing gradients during GANs training.

\subsection{Learning Instability and Nash Equilibrium}
The architectural characteristics of GAN involve a complex interplay between the two deep neural networks in an adversarial manner. Their training happens in a cooperative yet competitive way using a zero-sum game strategy where both $G$ and $D$ aim to optimize their respective objective functions to achieve the Nash equilibrium i.e., a state beyond which they can not improve their performance unilaterally \cite{nash1951non}. While this cooperative architecture aims to optimize a global loss function, the optimization problems faced by the individual networks are fundamentally opposing. Due to this complexity in the loss function, there can be situations where some minor adjustments in one network can trigger substantial modifications in the other. Moreover, when both the networks aim to independently optimize their loss functions without coordination, attaining the Nash equilibrium can be hard. Such instances of desynchronization between the networks can lead to instability in the overall learning process and substantially increase the computation time \cite{luo2018towards}. To counter this challenge, recent advancements in GAN architectures have been focusing on enhancing training stability. The feature matching technique improves the stability of the GAN framework by introducing an alternative cost function for $G$ combining the output of the discriminator \cite{salimans2016improved}. Additionally, historical averaging of the parameters \cite{salimans2016improved}, unrolled GAN \cite{metz2016unrolled}, and gradient penalty \cite{gulrajani2017improved} strategies mitigate learning instability and promote convergence of the model.

\subsection{Stopping Problem}

During GANs training, determining the appropriate time at which the networks are fully optimized is crucial for addressing the problems related to overfitting and underfitting. However, in GANs due to the minimax objective function determining the state of the networks based on their respective loss functions is impossible. To address this issue related to the GANs stopping criterion, researchers often employ an early stopping approach where the training halts based on a predefined threshold or the lack of improvement in evaluation metrics.

\subsection{Internal Distributional Shift}
The internal distributional shift often called internal covariate shift refers to the changing distribution in the network activations of the current layer w.r.t the previous layer. In the context of GAN, when the generator's parameters are updated, the distribution of its output may change, leading to internal distributional shifts in subsequent layers and causing the discriminator's learning to lag behind. This phenomenon affects the convergence of the GAN training process and the computational complexity of the network significantly increases to counter the shifts. To address this issue batch normalization technique is widely adopted in various applications of GAN \cite{ioffe2015batch}.

\section{DISCUSSION}\label{Section_Discussion}

Over the past decade, GANs have emerged as the foremost and pivotal generative architecture within the areas of computer vision, natural language processing, and related fields. To enhance the performance of GAN architecture, numerous studies have focused on the following: (i) the generation of high-quality samples, (ii) diversity in the simulated samples, and (iii) stabilizing the training algorithm. Constant efforts and improvements of the GAN model have resulted in plausible sample generation, text/image-to-image translations, data augmentation, style transfer, anomaly detection, and other applied domains.

Recent advancements in machine learning with the help of Diffusion models \cite{sohl2015deep, ho2020denoising, song2019generative} also known as score-based generative models have made a strong impression on a variety of tasks including image denoising, image inpainting, image super-resolution, and image generation. The primary goal of Diffusion models is to learn the latent structure of the dataset by modeling the way in which data points diffuse through the latent space. \cite{dhariwal2021diffusion} has shown that Diffusion models outperform GANs on image synthesis due to their better stability and non-existence of mode collapse. However, the cost of synthesizing new samples and computational time for making realistic images lead to its shortcomings when applied to real-time application \cite{croitoru2023diffusion, saharia2022palette}. Due to the fact that GANs need fine-tuning in their hyperparameters, Transformers \cite{vaswani2017attention} have been used to enhance the results of GANs that can adopt self-attention layers. This helps in designing larger models and replacing the neural network models of $G$ and $D$ within the GAN structure. TransGAN \cite{jiang2021transgan} introduces a GAN architecture without convolutions by using Transformers in both $G$ and $D$ of the GAN resulting in improved high-resolution image generation. \cite{lv2022improved} presented an intersection of GANs and Transformers to predict pedestrian paths. Although Transformers and their variants have several advantages, they suffer from high computational (time and resource) complexity \cite{sasal2022w}. More recently, physics-informed neural networks (PINN) \cite{raissi2019physics} was introduced as a universal function approximator that can incorporate knowledge of physical laws to govern the data in the learning process. PINNs overcome the low data availability issue \cite{elabid2022knowledge} in which GANs and Transformers lack robustness, rendering them ineffective scenarios. A GAN framework based on a physics-informed (PI) discriminator for uncertainty quantification is used to inform the knowledge of physics during the learning of both $G$ and $D$ models. Physics-informed Discriminator GAN (PID-GAN) \cite{daw2021pid} doesn't suffer from an imbalance of generator gradient from multiple losses. Another architecture namely Physics-informed GAN (PI-GAN) \cite{yang2021measure} tackles the problem of sequence generation with limited data. It integrates a transition module in the generator part that can iteratively construct the sequence with only one initial point as input. Solving differential equations using GANs to learn the loss function was presented in the Differential Equation GAN (DEQ-GAN) model \cite{bullwinkel2022deqgan}. Combining GANs with PINNs achieved solution accuracies that are competitive with popularly used numerical methods. 

Large language models (LLMs) \cite{radford2019better} became a very popular choice for their ability to understand and generate human language. LLMs are neural networks that are trained on massive text datasets to understand the relationship between words and phrases. This enables LLMs to generate text that is both coherent and grammatically correct. Recently, LLMs and their cousin ChatGPT revolutionized the field of natural language processing, question-answering, and creative writing. Additionally, LLMs and their variants are used to create creative content such as poems, scripts, and codes. GANs and LLMs are two powerful co-existing models where the former is used to generate realistic images. Mega-TTS \cite{jiang2023mega} adopt a VQGAN \cite{esser2021taming} based acoustic model and a latent-code language model called Prosody-LLM (P-LLM) \cite{ren2022prosospeech} to solve zero-shot text-to-speech at scale with intrinsic inductive bias. Future works in the hybridization of GANs with several other architectures will be a promising field of future research.

\section{FUTURE RESEARCH DIRECTION}\label{Section_Future}

Despite the substantial advancements achieved by GAN-based frameworks over the past decade, there remain a number of challenges spanning both theoretical and practical aspects that require further exploration in future research. In this section, we identify these gaps that necessitate deeper investigation to enhance our comprehension of GANs. The summary is presented below:

\paragraph{Fundamental questions on the theory of GANs} Recent advancements in the theory of GAN by \cite{liu2017approximation, biau2020some, biau2021some} explored the role of the discriminator family in terms of JS divergence and some large sample properties (convergence and asymptotic normality) of the parameter describing the empirically selected generator. However, a fundamental question of how well GANs can approximate the target distribution $p^*$ remained largely unanswered. From the theoretical perspective, there is still a mystery about the role and impact of the discriminator on the quality of the approximation. The universal consistency and the rate of convergence of GANs and their variants still remain an open problem.
    
\paragraph{Improvement of training stability and diversity} Achieving the Nash equilibrium in GAN frameworks, which is essential for the generator to learn the actual sample distribution, requires stable training mechanisms \cite{ratliff2013characterization, arora2017gans}. However, attaining this optimal balance between the generator and discriminator remains challenging. Various approaches have been explored, such as WGAN \cite{arjovsky2017wasserstein}, SN-GAN \cite{miyato2018spectral}, One-sided Label Smoothing \cite{szegedy2016rethinking}, and WGAN with gradient penalty (WGAN-GP) \cite{gulrajani2017improved}, to enhance training stability. Additionally, addressing mode collapse, a common GAN issue that leads to limited sample diversity, has prompted strategies like WGAN \cite{arjovsky2017wasserstein}, U-GAN \cite{metz2016unrolled}, generator regulating GAN (GRGAN) \cite{wang2020multimodal}, and Adaptive GAN \cite{tolstikhin2017adagan}. Future research could focus on devising techniques to stabilize GAN training and alleviate problems like mode collapse through regularization methods, alternative loss functions, and optimized hyperparameters. Incorporating methods like multi-modal GANs, designed to generate diverse outputs from a single input, might contribute to enhancing sample diversity \cite{wang2020multimodal}.

\paragraph{Data scarcity in GAN} Addressing the issue of data scarcity in GANs stands as a crucial research trajectory. To expand GAN applications, forthcoming investigations could focus on devising training strategies for scenarios with limited data. Approaches such as few-shot GANs, transfer learning, and domain adaptation offer the potential to enhance GAN performance when data is scarce \cite{hariharan2011semantic, tzeng2017adversarial}. This challenge becomes especially pertinent when acquiring substantial datasets poses difficulties. Additionally, refining training algorithms for maximal data utility could be pursued. Bolstering GAN effectiveness in low-data situations holds pivotal significance for broader adoption across various industries and domains.

\paragraph{Ethics and privacy}Since its inception in 2014, GAN development has yielded substantial benefits in research and real-world applications. However, the inappropriate utilization of GANs can give rise to latent societal issues such as producing deceptive content, malicious images, fabricated news, deepfakes, prejudiced portrayals, and compromising individual safety \cite{afchar2018mesonet}. To tackle these issues, the establishment of ethical guidelines and regulations is imperative \cite{taeihagh2021governance}. Future research avenues might center on developing robust techniques to detect and alleviate ethical concerns associated with GANs, while also advocating their ethical and responsible deployment in diverse fields. Essential to this effort is the creation of forgery detection methods capable of effectively identifying AI-generated content, including images produced through GANs. Furthermore, GANs can be susceptible to adversarial attacks, wherein minor modifications to input data result in visually convincing yet incorrect outputs \cite{liu2022distributed, lucic2018gans}. Future investigations could prioritize the development of robust GANs that can withstand such attacks, alongside methods for identifying and countering them. Ensuring the integrity and reliability of GANs is of utmost importance, particularly in contexts like authentication, content verification, and cybersecurity \cite{goodfellow2014explaining, samangouei2018defense}.

\paragraph{Real-time implementation and scalability} While GANs have shown immense potential, their resource-intensive nature hinders real-time usage and scalability. Recent GAN variants like ProGAN \cite{karras2017progressive} and Att-GAN \cite{he2019attgan} aim to address this complexity. Future efforts might focus on crafting efficient GAN architectures capable of generating high-quality samples in real-time, vital for constrained platforms like mobile devices and edge computing. Integrating GANs with reinforcement learning, transfer learning, and supervised learning, as seen in RidgeGAN \cite{thottolil2023prediction}, opens opportunities for hybrid models with expanded capabilities. Research should delve into hybrid approaches, leveraging GANs alongside other techniques for enhanced generative potential. Additionally, exploring multimodal GANs that produce diverse outputs from multiple modalities can unlock novel avenues for creating complex data \cite{hausknecht2015deep}.

\paragraph{Human-centric GANs} GANs have the potential to enable human-machine creative cooperation \cite{yang2017lr}. Future research could emphasize human-centric GANs, integrating human feedback, preferences, and creativity into the generative process. This direction might pave the way for interactive and co-creative GANs, enabling the production of outputs aligned with human preferences and needs, while also involving users in active participation during the generation process.

\paragraph{Other innovative applications and industry usage} 
Initially designed for generating realistic images, GANs have exhibited impressive performance in computer vision. While their application has extended to domains like time series generation \cite{yang2023cnts, li2023m3gan}, audio synthesis \cite{yang2017midinet}, and autonomous vehicles \cite{ghosh2016sad}, their use outside computer vision remains somewhat constrained. The divergent nature of image and non-image data introduces challenges, particularly in non-image contexts like NLP, where discrete values such as words and characters predominate \cite{alvarez2022gans}. Future research can aim to overcome these challenges and enhance GANs' capabilities in discrete data scenarios. Furthermore, exploring unique applications of GANs in fields like finance, education, and entertainment offers the potential to introduce new possibilities and positively impact various industries \cite{antipov2017face}. Collaborative efforts across disciplines could also harness diverse expertise, fostering synergies to enhance GANs' adaptability across a broad spectrum of applications \cite{mohamed2016learning}.

\section{CONCLUSION}\label{Section_Conclusion}
In this article, we presented a GAN survey, GAN variants, and a detailed analysis of the wide range of GAN applications in several applied domains. In addition, we reviewed the recent theoretical developments in the GAN literature and the most common evaluation metrics. Despite all these one of the core contributions of this survey is to discuss several obstacles of various GAN architectures and their potential solutions for future research. Overall, we discuss GANs' potential to facilitate practical applications not only in image, audio, and text but also in relatively uncommon areas such as time series analysis, geospatial data analysis, and imbalanced learning. In the discussion section, apart from GANs' significant success, we detail the failures of GANs due to their time complexity and unstable training. Although GANs have been phenomenal for the generation of hyper-realistic data, current progress in deep learning depicts an alternative narrative. Recently developed architectures such as Diffusion models have demonstrated significant success and outperformed GANs on image synthesis. On the other hand, Transformers, a deep learning architecture based on a multi-head attention mechanism, has been used within GAN architecture to enhance its performance. Furthermore, Large Language Models, a widely utilized deep learning structure designed for comprehending and producing natural language, have been incorporated into GAN architecture to bolster its effectiveness. The hybridization of PINN and GAN namely, PI-GAN can solve inverse and mixed stochastic problems based on a limited number of scattered measurements. On the contrary, GANs' ability which relies on large data for training, using physical laws inside GANs in the form of stochastic differential equations can mitigate the limited data problem. Several hybrid approaches combining GAN with other powerful deep learners are showing great merit and success as discussed in the discussion section. Finally, several applications of GANs over the last decade are summarized and criticized throughout the article. 

\bibliographystyle{IEEEtran}
\bibliography{IEEEabrv,Bibliography}

\end{document}